\documentclass{article}

\usepackage[nonatbib,preprint]{neurips_2023}

\usepackage[utf8]{inputenc} 
\usepackage[T1]{fontenc}    
\usepackage{hyperref}       
\usepackage{url}            
\usepackage{booktabs}       
\usepackage{amsfonts}       
\usepackage{nicefrac}       
\usepackage{microtype}      
\usepackage{xcolor}         

\usepackage{graphicx}
\usepackage{amsmath}
\usepackage{amssymb}
\usepackage{mathtools}
\usepackage{amsthm}

\usepackage[capitalize,noabbrev]{cleveref}

\usepackage[shortlabels]{enumitem}
\setlist{nosep}
\usepackage{tikz-cd,extarrows}
\usetikzlibrary{decorations.pathmorphing}
\usepackage{subcaption}
\usepackage{listings}
\definecolor{mygreen}{rgb}{0,0.6,0}
\definecolor{mygray}{rgb}{0.5,0.5,0.5}
\definecolor{mymauve}{rgb}{0.58,0,0.82}
\theoremstyle{plain}

\theoremstyle{definition}

\theoremstyle{remark}

\newcommand{\nrm}[1]{\lvert #1 \rvert}

\newcommand{\cD}{\mathcal{D}}

\DeclareMathOperator{\relu}{ReLU}

\DeclareMathOperator{\Acc}{Acc}

\usepackage[style=alphabetic,sortcites]{biblatex}
\title{Edit at your own risk: evaluating the robustness of edited models to distribution shifts}

\author{%
  Davis Brown\(^{1\thanks{Equal contribution}}\) 
  \And
  Charles Godfrey\(^{1\footnotemark[1]}\)
  \And
  Cody Nizinski\(^{1}\)
  \And
  Jonathan Tu\(^{1}\)
  \And
  Henry Kvinge\(^{1, 2}\)
  \AND
  \\
  \(^1\)Pacific Northwest National Laboratory \(^2\)University of Washington
\AND
\texttt{\{first\}.\{last\}@pnnl.gov}
}

\begin{document}

\maketitle
\begin{abstract}
The current trend toward ever-larger models makes standard retraining procedures an ever-more expensive burden.
For this reason, there is growing interest in \emph{model editing}, which enables computationally inexpensive, interpretable, post-hoc model modifications. 
While many model editing techniques are promising, research on the properties of edited models is largely limited to evaluation of validation accuracy.
The robustness of edited models is an important and yet mostly unexplored topic.
In this paper, we employ recently developed techniques from the field of deep learning robustness to investigate both how model editing affects the general robustness of a model, as well as the robustness of the specific behavior targeted by the edit.
We find that edits tend to reduce general robustness, but that the degree of degradation depends on the editing algorithm and layers chosen.
Motivated by these observations we introduce a new model editing algorithm, {\emph{1-layer interpolation (1-LI)}}, which uses weight-space interpolation to navigate the trade-off between editing task accuracy and general robustness.
\end{abstract}

\section{Introduction}



Not only is the initial process of training a large model expensive and time consuming, but so are any additional rounds of fine-tuning. 
Since many applications require a model to be repeatedly modified after initial training to address model deficiencies and other undesirable behavior, efficient and lightweight methods for modifying a model on-the-fly are of widespread interest. 

One family of such  methods is called {\emph{model editing}}. These methods aim to modify a model without the need for extensive computations
or a large collection of new training examples. Most of these methods aim to
change the way a model behaves in very specific instances, for example modifying
an image model so that it correctly classifies cars even on streets covered
with snow \cite{santurkarEditingClassifierRewriting2021a} or updating a language
model to correctly identify the new prime minister of the United Kingdom \cite{mitchellFastModelEditing2022,mitchellMemoryBasedModelEditing2022}.

In the usual framework, the goal of model editing has been to modify the model so as to (1) maintain performance on the original base dataset while (2) increasing performance for the editing dataset or task. However, since almost all real-world applications requre some amount of generalization to shifts in distribution, even if a model edit achieves (1) and (2), it is essential to explore how edits effect the robustness of the model outside of test set performance. In fact, it has become painfully apparent that high validation accuracy is a myopic measure of model performance (see for example \S 2 of \cite{hendrycksUnsolvedProblemsML2022}). We therefore ask
\begin{center}
    {\em To what extent do edited models generalize to natural distribution shifts?}
\end{center}
Given that many editing methods are quite different from traditional network optimization and fine-tuning, the answer to this question is not obvious. Indeed, one could imagine that highly targeted (and possibly clumsy) modifications could either overfit the model to the editing task or generally degrade some of the highly-tuned feature extraction abilities that allow for generalization. Given this motivation our investigation takes two parts.

\textbf{An empirical study of editing robustness using an evaluation framework that takes the editing objectives into account}: Many works have shown deep learning models to be more brittle than previously recognized, suffering large drops in accuracy when presented with subtle distribution shifts in input data \cite{rechtImageNetClassifiersGeneralize2019,hendrycksBenchmarkingNeuralNetwork2019,barbuObjectNetLargescaleBiascontrolled2019,hendrycksNaturalAdversarialExamples2021,hendrycks2021many}. We aim to perform a similar study focused on editing methods. This means editing a range of models using different editing methods and exploring resulting model robustness. We note that because the objective in editing is more complex than the objective for a generic classifier (accuracy on the test set must be preserved while increasing performance on the editing task), there are more ways for a model to lack robustness. Consequently, a new evaluation framework suitable for edited models is needed. We describe this in \Cref{sec:ood}. We then use datasets that simulate types of distribution shift found in the real world to draw conclusions about the robustness penalty one incurs by editing a model. Beyond robustness evaluation, we hope our experimental framework will prove a useful template for other kinds of explorations of the effects of editing on model performance.

\begin{figure}
  \vskip -0.2in
\begin{center}
\begin{subfigure}[b]{0.95\textwidth}
   \includegraphics[width=1\linewidth]{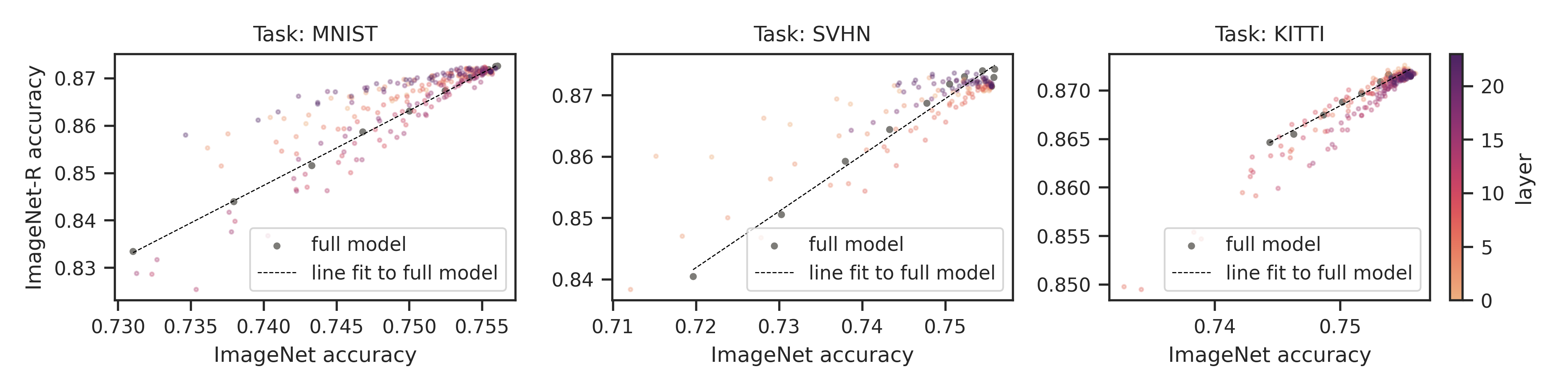}
\end{subfigure}
\begin{subfigure}[b]{0.95\textwidth}
   \includegraphics[width=1\linewidth]{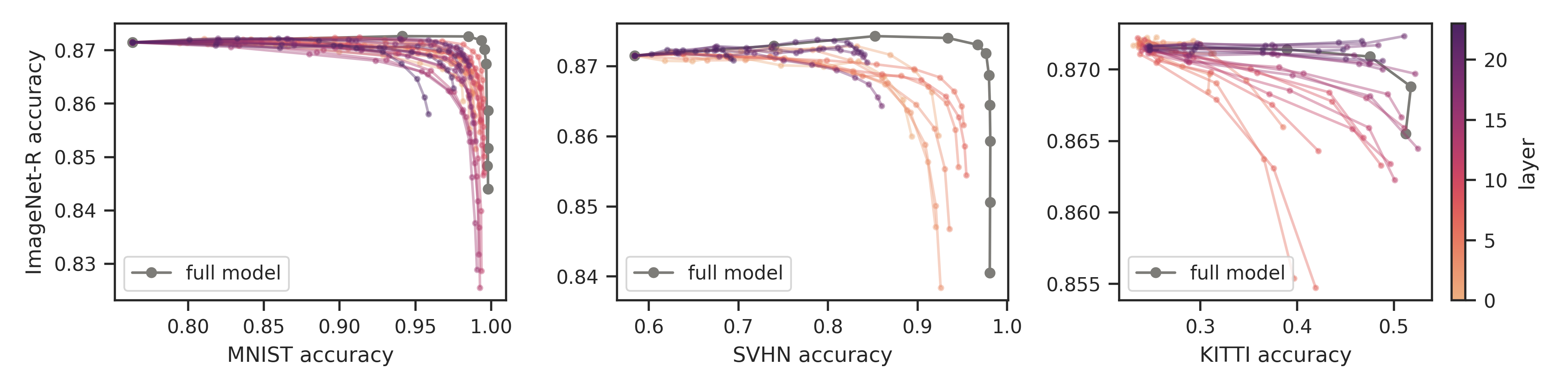}
  \label{fig:1-L-imagenetR}
\end{subfigure}
\caption[Effective robustness]{1-layer interpolation displays stand-out \textit{effective robustness} after fine-tuning on an editing task. See Section \ref{sec:results-1li}. ImageNet-R vs ImageNet (\textbf{top}) and task accuracy (\textbf{bottom}) along line segments \((1-\alpha) W + \alpha W' \) between original and edited weights \(W \) and \(W'\) of CLIP models edited on various tasks, where different points correspond to different interpolation parameters \(\alpha \).}   \label{fig:INR-scatter}
 \end{center}
   \vskip -0.2in
\end{figure}

\textbf{Preserving robustness via weight interpolation}: With benchmarks for model degradation in hand, we ask whether we can mitigate the robustness penalties incurred because of editing. To probe this question we leverage a recently observed phenomenon whereby \emph{linearly} interpolating between pretrained weights \(W \) and weights \(W' \) fine-tuned on a separate task results in a model that can largely solve both the pretraining and fine-tuning tasks (e.g., \cite{ilharcoPatchingOpenvocabularyModels2022}).\footnote{This phenomenon is perhaps unexpected considering the highly non-convex nature of the objective functions involved.}
Our results show that despite the large difference in optimization methods between editing and traditional deep learning training where this phenomenon has been observed \cite{ilharcoPatchingOpenvocabularyModels2022}, one can effectively balance original task performance, edited task performance, and model robustness via weight interpolation. These results can be viewed as new observations of the monotone linear interpolation phenomenon.
We highlight {\emph{1-layer interpolation (1-LI)}}, a novel and particularly well-performing variant of weight interpolation.
The 1-LI method consists of single-layer fine-tuning followed by weight space interpolation, and results in less robustness degradation than whole-model fine-tuning and interpolation (\Cref{fig:INR-scatter}), while remaining highly effective on the target task. 


In summary, our \noindent\textbf{contributions:} 
\begin{itemize}[nosep]
\item We provide, to our knowledge, the most comprehensive benchmark evaluations to-date of model editing techniques in order to understand their effect on robustness. Our framework examines performance and provides robustness metrics for both original and updated editing tasks.
\item While in general editing tends to reduce robustness, our results identify editing algorithms that suffer less degradation than their counterparts. For example, editing different layers on an editing task can result in similar accuracy but different robustness to distribution shift.
\item We introduce new approaches to editing, most notably \emph{1-layer interpolation}, which results in less robustness degradation than previous weight-space interpolation methods and displays the phenomenon of \textit{effective robustness} exhibited in \Cref{fig:1-L-imagenetR}.
\end{itemize}

\section{Related work}
\label{sec:rw}



\noindent\textbf{Model robustness and generalization:} 
Robustness to distribution shift is an important topic in machine learning, in large part because it addresses a key challenge in real-world model deployment.
Pioneering studies have revealed that even when models achieve superhuman performance on a held-out test set, their performance often degrades significantly on alternative test sets that are either mildly corrupted \cite{hendrycksBenchmarkingNeuralNetwork2019} or out of distribution
\cite{rechtImageNetClassifiersGeneralize2019}. Many of these works introduced datasets or tasks designed to evaluate various aspects of model robustness \cite{wilds2021, sagawa2022extending, hendrycksBenchmarkingNeuralNetwork2019,hendrycks2021many,geirhos2018imagenet}.
We exploit several of these techniques and tools to investigate the robustness of model editing, which has not previously been studied.  


\noindent\textbf{Editing methods:} 
The concept of model editing was first introduced in the context of generative adversarial networks \cite{bauRewritingDeepGenerative2020}.
Model edits were implemented by idealizing a neural network layer as a linear associative memory and then, based on that idealization, deriving a low-rank update to the layer weights based on an input or collection of inptuts (a `concept').
That approach has also been employed to alter the behavior of image classifiers \cite{santurkarEditingClassifierRewriting2021a} and autoregressive language models \cite{mengLocatingEditingFactual2022,mengMassEditingMemoryTransformer2022}. 
In addition, a variety of alternative editing algorithms have been proposed.
\cite{decaoEditingFactualKnowledge2021} and \cite{mitchellFastModelEditing2022} trained hyper-networks to update model weights.
\cite{daiKnowledgeNeuronsPretrained2022} identified ``knowledge neurons'' and then surgically modified them.
\cite{mitchellMemoryBasedModelEditing2022} augmented a target model with an auxiliary detector trained to identify inputs relevant to the editing task, and then diverted them to an edited model. See also \cite{hase2021language} for analysis of beliefs stored in language models and methods to update those beliefs.
Recent work also extended model editing to open vocabulary image-language models (e.g., CLIP), where it was demonstrated that interpolating between the original and edited weights allows users to navigate a trade-off between performance on the original and edited tasks \cite{ilharcoPatchingOpenvocabularyModels2022}. 

Of particular relevance to our work are recent findings of \cite{lee2023surgical}, which found that updating only a subset of model weights can result in greater robustness to distribution shift. We consider a different notion of robustness in this work, as well as a different set of models and editing/fine-tuning tasks than that work. We likewise update a more granular subset of weights (\cite{lee2023surgical} updates "blocks" of weights, whereas we typically update weights in a single layer).

\section{Edit tasks and data}
\label{sec:tasks-data}




The model editing problem makes use of a neural network \(f\)  trained with a standard neural network optimization procedure 
and evaluated on an \emph{original task} specified by a dataset \(\cD_\text{orig}^\text{val}\). In our experiments we study image classification and open-vocabulary (namely, CLIP) models evaluated on ImageNet \cite{imagenet_cvpr09} object recognition. The \emph{editing task} is defined by a different (typically much smaller) dataset \(\cD_{\text{edit}}^{\text{train}}\).
We consider two analogous families of editing tasks. 
For both types of editing tasks, there is a novel editing validation dataset \(\cD_{\text{edit}}^{\text{val}}\) consisting of input-output pairs \((x, y)\).

\subsection{First group of editing tasks (from \cite{santurkarEditingClassifierRewriting2021a})}
\label{sec:santurkar-data}
The first tasks are related to experiments performed in \cite{santurkarEditingClassifierRewriting2021a} with the goal of modifying model behavior for specific input  \textit{concepts}. 
We make use of two datasets from \cite{santurkarEditingClassifierRewriting2021a}: a ``vehicles on snow'' dataset and a large-scale synthetic dataset.

\textbf{``Vehicles on snow'' dataset}. Motivated by the observation that an image classifier could accurately identify vehicles on roadways, but was noticeably less accurate when classifying vehicles on snow, \cite{santurkarEditingClassifierRewriting2021a} took datapoints \((x,y ) \) from ImageNet and used style transfer to replace road segments in the image \(x\) with snow texture, yielding a a modified image \(  x' \).
The result is a training dataset \(\cD_{\text{edit}}^{\text{train}}\) of triples \((x, x', y) \) which can be used in any of the editing algorithms described below. 
Validation is performed using a curated collection of images of vehicles on show downloaded from Flickr (the original source of ImageNet images). 
\\
\textbf{Synthetic concept-style datasets}. The synthetic pipeline creates a large number of datasets similar in spirit to the vehicles on snow dataset described above, as follows: given a `concept’ (e.g., “mountain”) and a style (e.g. "fall colors"), a pretrained segmentation model is used to identify the concept in an ImageNet datapoint $(x, y)$. Next, a style transfer model is used to generate a modified image $x'$ by transfering the style onto the concept (e.g. "mountain'' texture is transformed to ``fall colors'' texture). As in the case of ``vehicles on snow,'' the synthetic pipeline yields a triple $(x, x', y)$ for each concept-style pair. As in \cite{santurkarEditingClassifierRewriting2021a}, we select images for modification from the ImageNet validation set, then split the resulting image pairs into training set \(\cD_{\text{edit}}^{\text{train}}\) and validation set \(\cD_{\text{edit}}^{\text{val}}\) for the editing task.  Further details as well as example datapoints can be found in \cref{sec:corr-lss}. 

\subsection{Second group of editing tasks from \cite{ilharcoPatchingOpenvocabularyModels2022}}
The second editing tasks borrow the set-up from \cite{ilharcoPatchingOpenvocabularyModels2022}, where the goal is to edit the model to improve performance on certain \textit{tasks}.
For this benchmark, open-vocabulary models such as CLIP \cite{Radford2021LearningTV} are adapted to the MNIST \cite{mnist}, SVHN \cite{svhn} and KITTI (distance estimation) \cite{Geiger2013IJRR}. While traditionally viewed as a sort of fine-tuning or continual learning problem, we show that it can be approached using model editing methods such as single-layer fine-tuning, and that measuring the the robustness of the resulting edited models yields interesting results (see \cref{sec:1lirob,sec:newmli}). The interest in MNIST, SVHN and KITTI stems from the fact that for all its impressive zero-shot performance on a wide variety of datasets, the original CLIP model struggled on these three (famously underperforming logistic regression on raw pixels in the case of MNIST).

\section{Editing methods}
\label{sec:edit-methods}

Here we provide brief descriptions focussing on new methods, leaving more thorough details as well as discussion of preexisting methods to \cref{sec:edmeth,sec:editing-method-details}. For a rapid summary of the editing algorithms appearing in our experiments please see \cref{tbl:methods}.

For ease of exposition, we describe editing methods for a simplified model (for discussion on the adaptations to more complicated architectures, e.g., with residual connections, see \cref{sec:editing-method-details}). Let $f$ be a neural network defined as the composition $f = f_{L} \circ f_{l-1} \circ \dots \circ f_2 \circ f_1$,
where $f_{l}(x) = \sigma(W_lx)$, with $W_l$ an \(n_{l} \times n_{l-1}\) matrix for each \(l = 1, \dots, L\) and \(\sigma\) a coordinate-wise non-linearity (e.g., \(\relu\)). For any such $l$, let \(f_{\leq l}\) denote the composition of the first \(l\) layers: $f_{l} \circ f_{l-1} \circ \dots \circ f_2 \circ f_1$.
Similarly, let \(f_{> l}\) denote the composition of the last $L-l$ layers $f_L \circ f_{L-1} \circ \dots \circ f_{l+1}$.

\textbf{Methods of updating weights:} Consider an input pair \((x, x')\), as in the vehicles on snow or large scale synthetic tasks (\cref{sec:santurkar-data}).
If \(f_{\leq l}(x) = f_{\leq l}(x')\) for any \(l\), it trivially follows that \(f(x) = f(x')\), as desired. 
This is the basic motivation for the following weight update procedures, all of which attempt to minimize the mean squared error between \(f_{\leq l}(x)\) and \( f_{\leq l}(x')\) creating the situation where \(f_{\leq l}(x) \approx f_{\leq l}(x')\)---we refer to this situation as \emph{output collision}. \emph{Local (resp. global) fine-tuning for output collision} optimizes \(W_l\) (resp. \(W_1, \dots, W_l\)) using stochastic gradient descent (SGD). \emph{Rewriting}
    \cite{bauRewritingDeepGenerative2020,santurkarEditingClassifierRewriting2021a,mengLocatingEditingFactual2022,mengMassEditingMemoryTransformer2022} performs a certain low rank update on \(W_l\) derived using the theory of linear associative memories (for details see \cref{sec:edmeth} and the references). 
For fine-tuning tasks where updates use pairs $(x',y)$ rather than $(x,x')$, additional methods are available, including local fine-tuning at layer $l$ and global fine-tuning from layer $l$ forward.

\textbf{Approaches to balancing original and editing accuracy}: The standard approach is cross validation, where one monitors accuracy on the original and editing validation datasets while performing optimization with respect to the editing task. One chooses a desirable ``best iteration'' (i.e. early stopping) as suits the application. Surprisingly, linearly interpolating weights often interpolates between performance on the original and editing tasks. Specifically, given original weights $W := (W_l)^L_{l = 1}$ and edited weights  $W' := (W_l')^L_{l = 1}$ one chooses $\alpha \in [0,1]$ such that $(1-\alpha)W + \alpha W'$ balances performance between the two tasks.
We note that weight space interpolations have only been explored for full-model fine-tuning. To our knowledge we are the first to observe that linear interpolation works when weights have been modified using an editing method that only updates weights at a single layer (see Section \ref{sec:results-1li}).

Finally, we introduce two new editing methods developed during our experiments on editing robustness.

\textbf{Single-layer interpolation (1-LI)}: We apply local fine-tuning at layer \(l\) along with weight-space interpolation. The inspiration for this approach was to combine a common feature of several editing algorithms, namely weight updates restricted to a single layer, with the empirical benefits of weight-space interpolation \cite{ilharcoPatchingOpenvocabularyModels2022, wortsmanRobustFinetuningZeroshot2022}. In Section \ref{sec:results-1li} we show that this method often reduces degradation of model robustness. 

\textbf{Direct low-rank editing}: This algorithm illustrates that some of the effectiveness of rewriting (and related methods mentioned in \cref{sec:rw}) stems \emph{simply from employing low-rank weight updates}. 
Implementation details, comparison with the methods of 
\cite{bauRewritingDeepGenerative2020,santurkarEditingClassifierRewriting2021a}, a code sample can be found in \cref{sec:editing-method-details}. 

\begin{table}[tb]
    \centering
    \begin{tabular}{p{4cm}lp{5cm}}
    \toprule
                                       method &                              figures &                                                                                                                                                                 reference \\
    \midrule
                        1-layer interpolation &                              3, 4, 5 &                                                                                                                                                                  \textbf{ours} \\
                      direct low-rank editing &                                 2(b) &                                                                                                                                                                  \textbf{ours} \\
       local fine-tuning for output collision &                                 2(a) & \textbf{ours}, but closely related to \cite{bauRewritingDeepGenerative2020,santurkarEditingClassifierRewriting2021a} \\
      global fine-tuning for output collision &                               1(a-b) &                                                              \textbf{ours}, but see also \cite{prior_learned_reps} \\
               local fine-tuning at layer $l$ &                                      &                                   general knowledge (in the context of robustness see \cite{lee2023surgical}) \\
                       full model fine-tuning & 1(a-b), $3^\ast$, $4^\ast$, $5^\ast$ &                     general knowledge ($^\ast$ combined with weight space interpolation \cite{ilharcoPatchingOpenvocabularyModels2022}) \\
                                    rewriting &                               1(a-c) &                                  \cite{bauRewritingDeepGenerative2020,santurkarEditingClassifierRewriting2021a} \\
    global fine-tuning from layer $l$ forward &                               1(a-b) &                                                                                                                                                         general knowledge \\
    \bottomrule
    \end{tabular}
    \caption{Editing methods studied in this paper, relevant figures and references where applicable.}
    \label{tbl:methods}
\end{table}

\section{Metrics for OOD performance of edited models}
\label{sec:ood}

We consider two different performance measures for an edited model on OOD data: the accuracy of an edited model on a distribution-shifted version of the original task validation set \(\cD_{\text{orig}}^{\text{val}}\), as well as the accuracy of the edited model on a distribution-shifted version of the \emph{editing} task validation set \(\cD_{\text{edit}}^{\text{val}}\). 
These two evaluations are quite different: in the example of the vehicles-on-snow task with a distribution shift generated by blurring images, the first performance measure requires evaluating the edited model on blurry images from all ImageNet classes (many of which are dogs), while the second requires evaluating blurry images of vehicles on snow. 
In other words, the former tests whether editing impacts the robustness of the model on the original task, and the latter tests the robustness of the edit itself. 


One common form of distribution shift for natural image data is \emph{corruption}. We model corrupted ImageNet images using the ImageNet-C dataset released by \cite{hendrycksBenchmarkingNeuralNetwork2019}, and use the ImageNet-C code to corrupt non-ImageNet data (see \cref{sec:corr-lss} for further details). ImageNet-C contains many forms of corruption at varying severity levels; a table as well as example images can be found in \cite{hendrycksBenchmarkingNeuralNetwork2019}. In addition, we evaluate models on ImageNet-R \cite{hendrycks2021many} (which contains paintings and drawings of ImageNet classes) and ImageNet-A \cite{hendrycksNaturalAdversarialExamples2021} (curated by downloading images related to ImageNet labels and retaining those incorrectly classified by a fixed ensemble of ResNet-50 models).

We use two ImageNet trained models: a VGG16 \cite{simonyanVeryDeepConvolutional2015} and a Vision Transformer (ViT) \cite{dosovitskiyImageWorth16x162021}. 
In addition, we experiment with an open-vocabulary CLIP model, using the ViT-L/14 weights available in the code release for \cite{Radford2021LearningTV}. This model was trained on a large dataset of 400 million image-text pairs scraped from the internet that is not publicly available. 


Let \(\cD_{\text{orig}}^{\text{val}} \) be a validation set for the original task, (for example, the ImageNet validation set) and let \(\tilde{\cD}_{\text{orig}}^{\text{val}} \) be a distribution shifted variant of \(\cD_{\text{orig}}^{\text{val}} \) (for example obtained by blurring images from \(\cD_{\text{orig}}^{\text{val}} \)). Similarly, let  \(\tilde{\cD}_{\text{edit}}^{\text{val}} \) be a distribution shifted variant of the editing validation set \(\cD_{\text{edit}}^{\text{val}} \). Letting \(f_{\text{orig}} \) and \(f_{\text{edited}}\) denote the model before and after editing,
we compute the following two metrics:
\begin{align}
    \textbf{original task OOD penalty: }&\Acc(f_{\text{edited}}, \tilde{\cD}_{\text{orig}}^{\text{val}} ) - \Acc(f_{\text{orig}}, \tilde{\cD}_{\text{orig}}^{\text{val}} ) \text{ and} \label{eq:orig-ood-penalty}\\
    \textbf{editing task OOD penalty: }&\Acc(f_{\text{edited}}, \tilde{\cD}_{\text{edit}}^{\text{val}} ) - \Acc(f_{\text{edit}}, \cD_{\text{edit}}^{\text{val}}). \label{eq:edit-ood-penalty}
\end{align}
In words, the original task OOD penalty is the raw change in accuracy on \(\tilde{\cD}_{\text{orig}}^{\text{val}} \) caused by model editing. On the other hand, the editing task OOD penalty is the decrease in editing validation performance caused by the distribution shift in question.



\section{Editing nearly always degrades model robustness \dots}\label{sec:rob-degrade}


Our first basic finding is that all editing methods result in \emph{negative} OOD penalties, meaning the edited models tend to be less robust. We give baseline editing performance in \Cref{fig:corr-lr-ft} (a), which displays editing validation accuracies of VGG and ViT models on the synthetic concept-style task. We take the maximum validation accuracy over the edited layers and averaging over all concept-style pairs considered.
The \textit{editing task OOD penalties} for the same models and editing tasks as \cref{fig:corr-lr-ft} (a) are given in \Cref{fig:corr-lr-ft} (b), where distribution shift is modelled by applying corruptions to the synthetic concept-style validation sets. As noted, all model edits considered result in penalties.

\Cref{fig:corr-lr-ft} (c) displays original task OOD penalties on corrupted ImageNet (i.e., corruptions of \(\cD_{\text{orig}}^{\text{val}}\)) for VGG16 models edited using the synthetic concept-style task, using the local fine-tuning for ouput collision and rewriting methods ((\ref{item:loc-ft-oc} and (\ref{item:rewrit} in Section \ref{sec:edit-methods}). Here we average over multiple factors of variation, including synthetic concept-style pair, editing layer and the different types of corruptions (blur, weather etc.) to display results with respect to corruption strength --- error bars represent the corresponding standard deviations.

\section{\dots but degradation depends on layer updated and editing method used}\label{sec:rob-depends}

Edits tend to degrade model robustness on the original \emph{and} editing task, but the magnitude of this decrease depends strongly on the method used. Broadly, methods constrained to only update a small fraction of model parameters can retain more robustness on the original task.

\begin{figure*}[t]
   \centering
   \begin{subfigure}{0.3\linewidth}
      \centering
      \includegraphics[width=\linewidth]{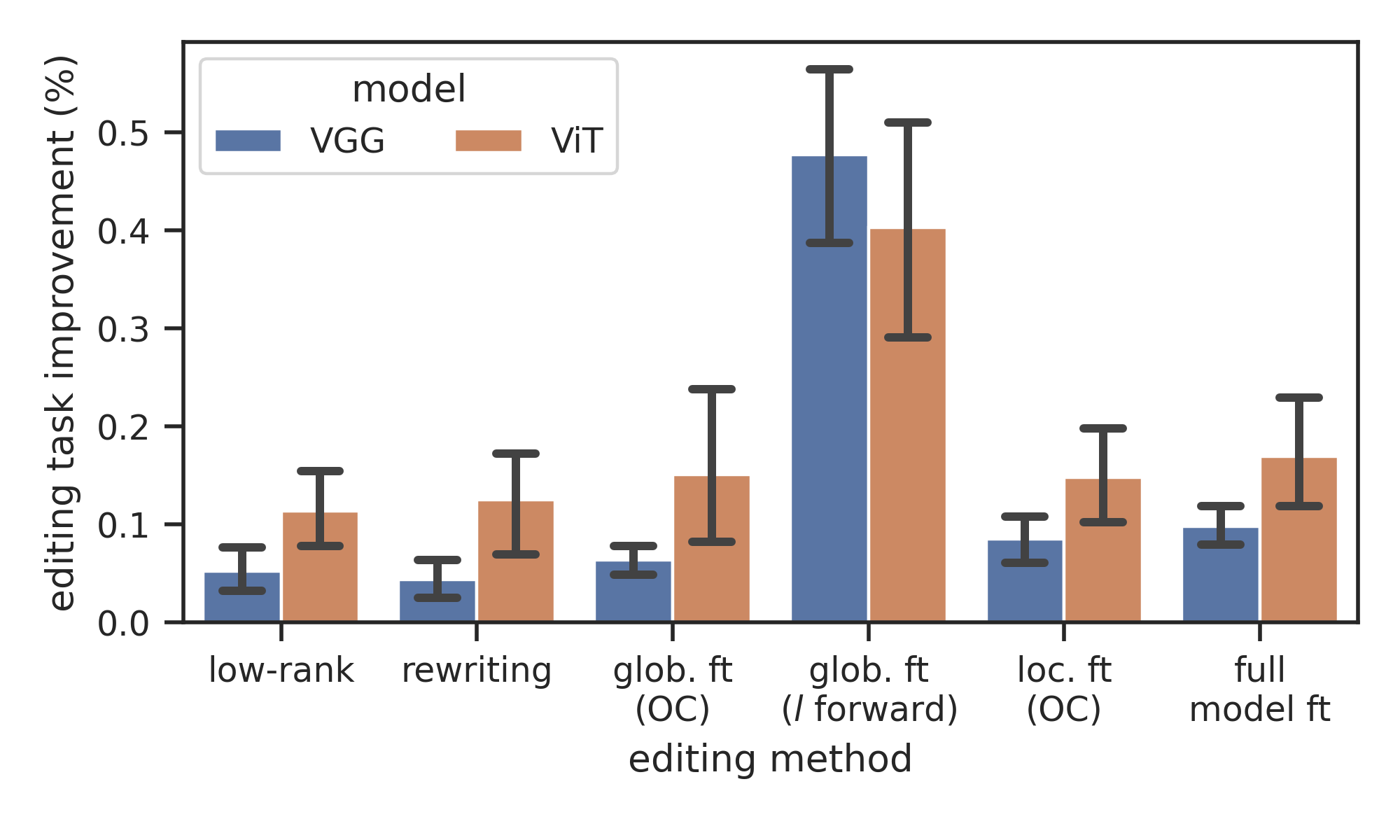}
      \caption{}
   \end{subfigure}
   \begin{subfigure}{0.3\linewidth}
      \centering
      \includegraphics[width=\linewidth]{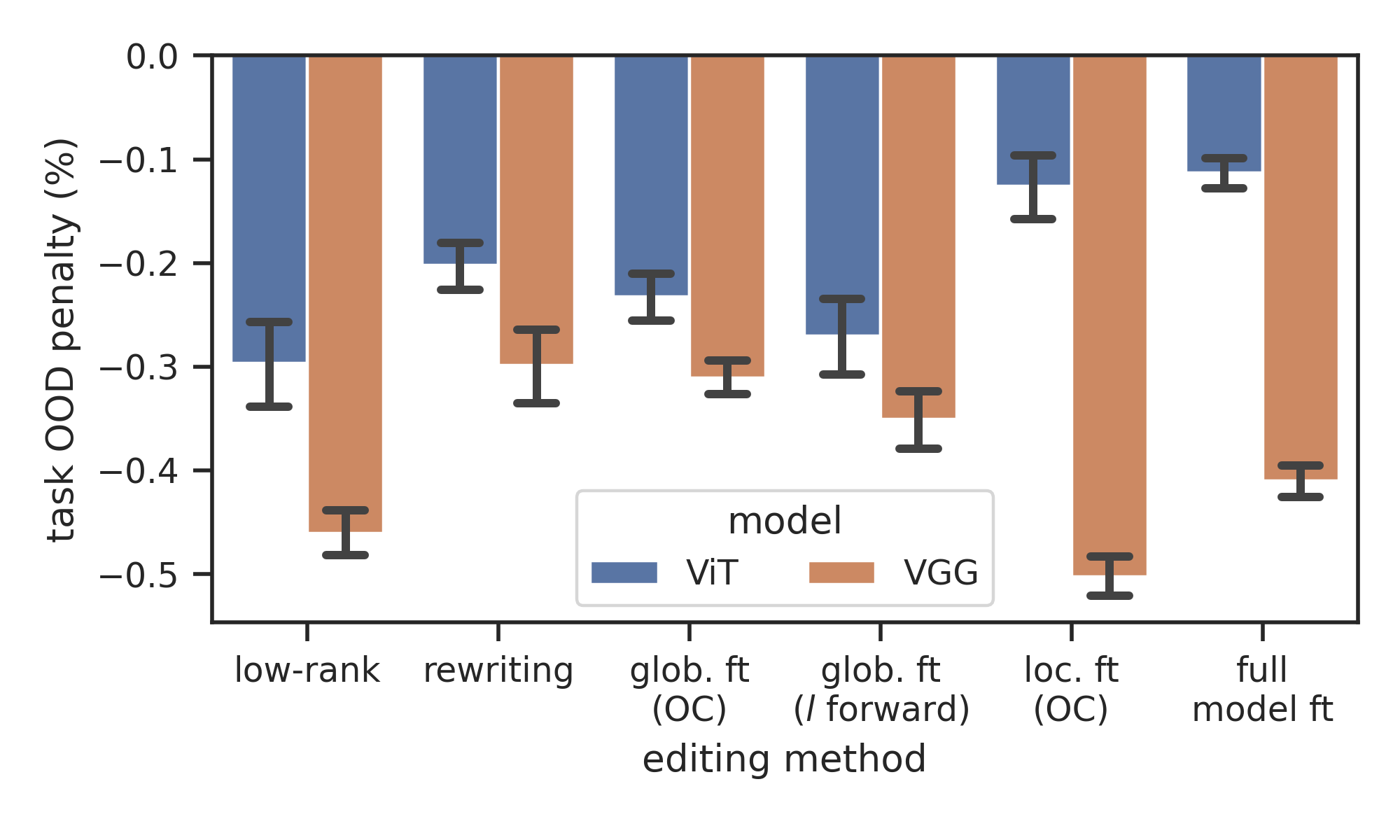}
      \caption{}
   \end{subfigure}
   \begin{subfigure}{0.3\linewidth}
      \centering
      \includegraphics[width=\linewidth]{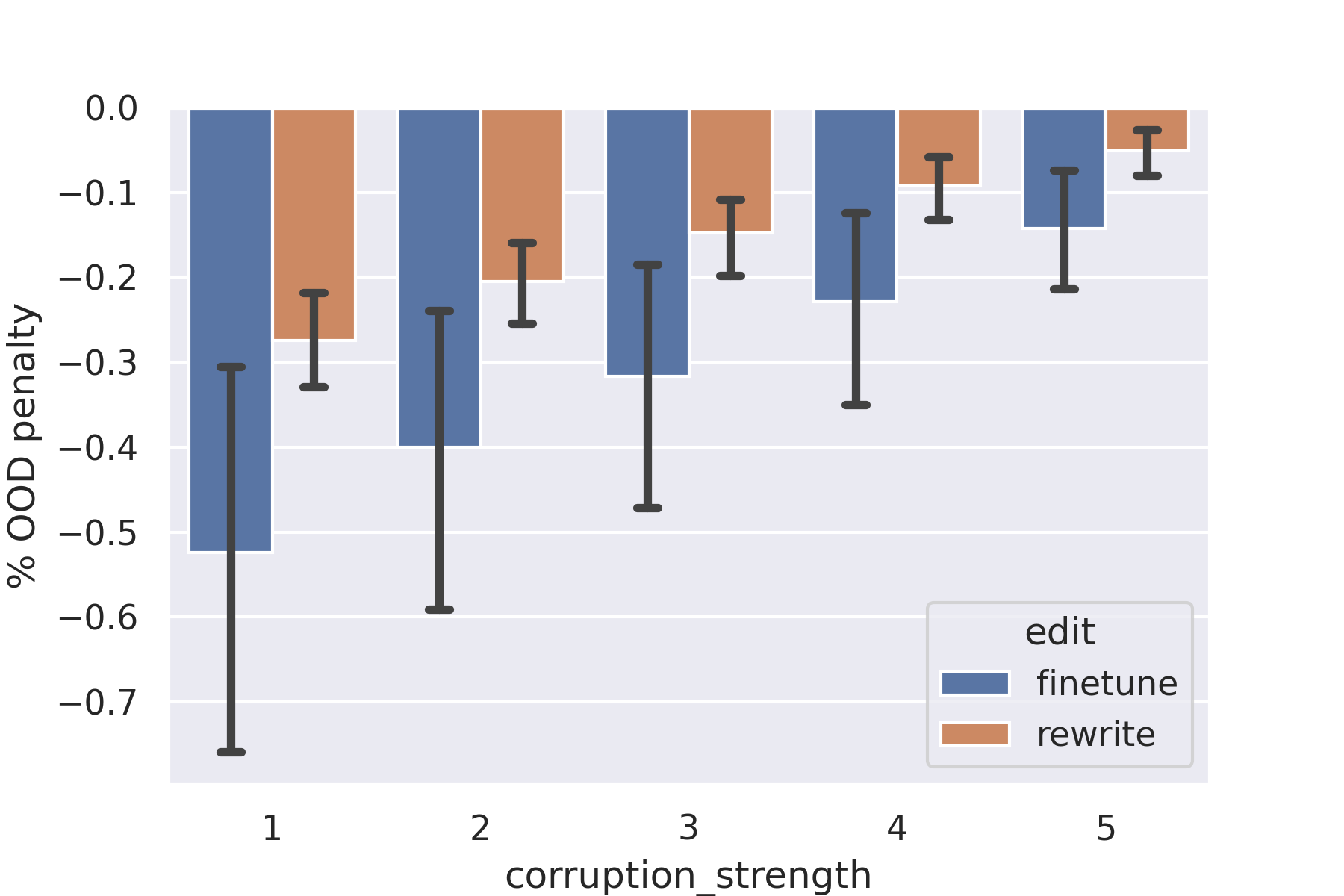}
      \caption{}
   \end{subfigure}
   \caption{\textbf{OOD performance of edited models}. \textbf{(left)} Editing validation accuracies of VGG16 and ViT models on the synthetic concept-style task. \textbf{(middle)} Editing task OOD penalties \cref{eq:edit-ood-penalty} of the edited models in (a). \textbf{(right)} original task OOD penalties on ImageNet-C \cref{eq:orig-ood-penalty} of VGG16 models edited using the synthetic concept-style task, using the local fine-tuning for ouput collision (\cref{item:loc-ft-oc}) and rewriting (\cref{item:rewrit}) methods. 
   }\label{fig:corr-lr-ft}
\end{figure*}

\textbf{Editing task penalties:} Observe that in \cref{fig:corr-lr-ft} (b) for both the VGG and ViT models the methods rewriting, global fine-tuning for output collison and global fine-tuning layers \(l \) forward  are ranked in increasing order of editing task OOD penalty. We find direct low-rank editing underperforms in terms of editing task OOD penalty (for both models) and that full model fine-tuning seems to lack a common trend for the two models.

\textbf{Original task penalties:} Models edited with rewriting incur lower original task OOD penalties at all levels of corruption strength, suggesting that they retain more of the robustness of the original, unedited model. 


\begin{figure*}[bt]
   \centering
   \begin{subfigure}{0.4\linewidth}
      \centering
      \includegraphics[width=\linewidth]{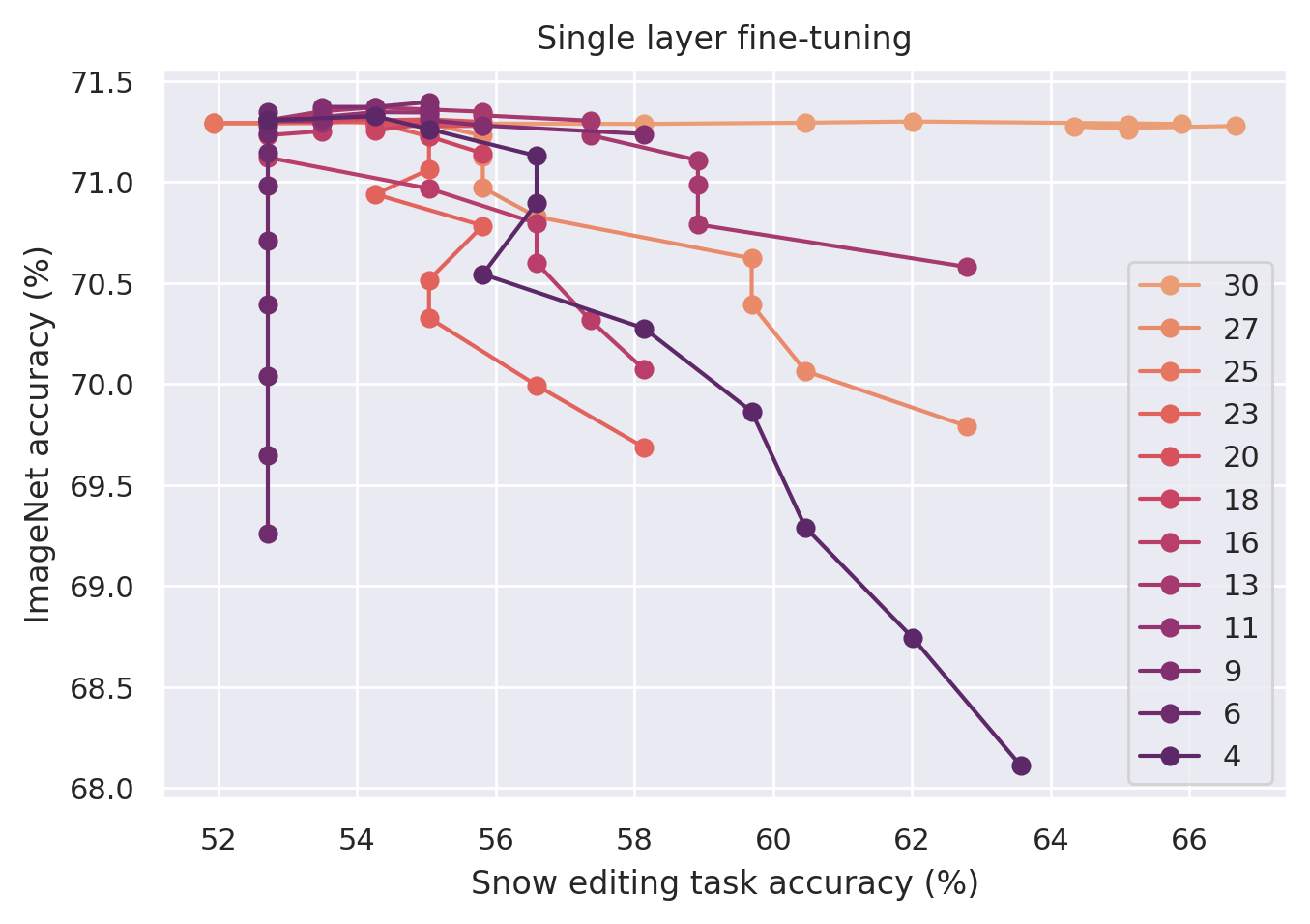}
      \caption{}
   \end{subfigure}
   \begin{subfigure}{0.4\linewidth}
      \centering
      \includegraphics[width=\linewidth]{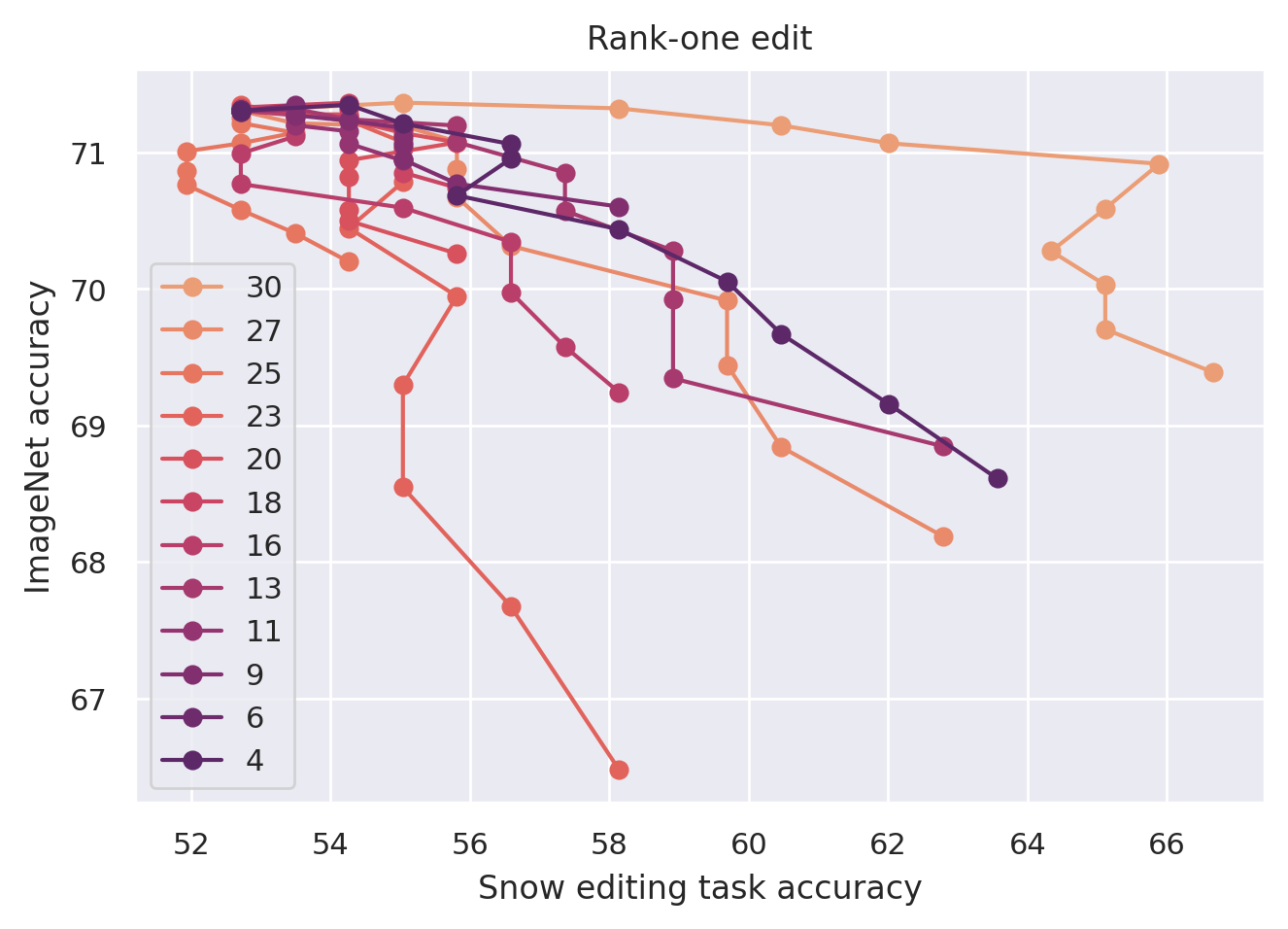}
      \caption{}
   \end{subfigure}
   \caption{Linear interpolation between original and edited weights \(W \) and \(W'\) for edited VGG16 models the vehicles-on-snow dataset. Each curve corresponds to a an editing layer, and each point on that curve corresponds to evaluation of a model with weights \((1-\alpha)W + \alpha W' \) for some \(t \in [0, 1] \). \textbf{(left)}: local fine-tuning for output collision. \textbf{(right)}: direct rank one editing.}\label{fig:single-layer-mli}
\end{figure*}

\Cref{fig:single-layer-mli} displays both ImageNet accuracy and vehicles-on-snow task accuracy along line segments \((1-\alpha)W + \alpha W'\) in weight space, where \(W \) are the original model weights and \(W' \) are the edited weights. In this experiment we use VGG16 models, and two different editing methods: local fine-tuning for output collision (\cref{item:loc-ft-oc}) and direct low-rank editing (with rank \(=1\)). For both editing methods we see that ImageNet accuracy generally decreases as editing accuracy increases, and moreover that the resulting curves appear to be roughly concave. This shows that weight space interpolation may serve as a viable alternative to early stopping for the purpose of balancing accuracy on the original and editing task. This was found to be the case for global fine-tuning in \cite{ilharcoPatchingOpenvocabularyModels2022}, but to our knowledge has not been established for local fine-tuning or low-rank editing. 

\section{Single layer interpolation (1-LI)}\label{sec:results-1li}

Weight interpolation continues to be an effective way of balancing original task accuracy and editing task accuracy, even when editing updates weights at a single layer (as opposed to global fine-tuning). Surprisingly, 1-LI methods outperform full fine-tuning and weight space interpolation at some robustness tasks.

\subsection{1-LI exhibits strong robustness}\label{sec:1lirob}
\begin{figure*}[tb]
  \centering
  \includegraphics[width=0.98\linewidth]{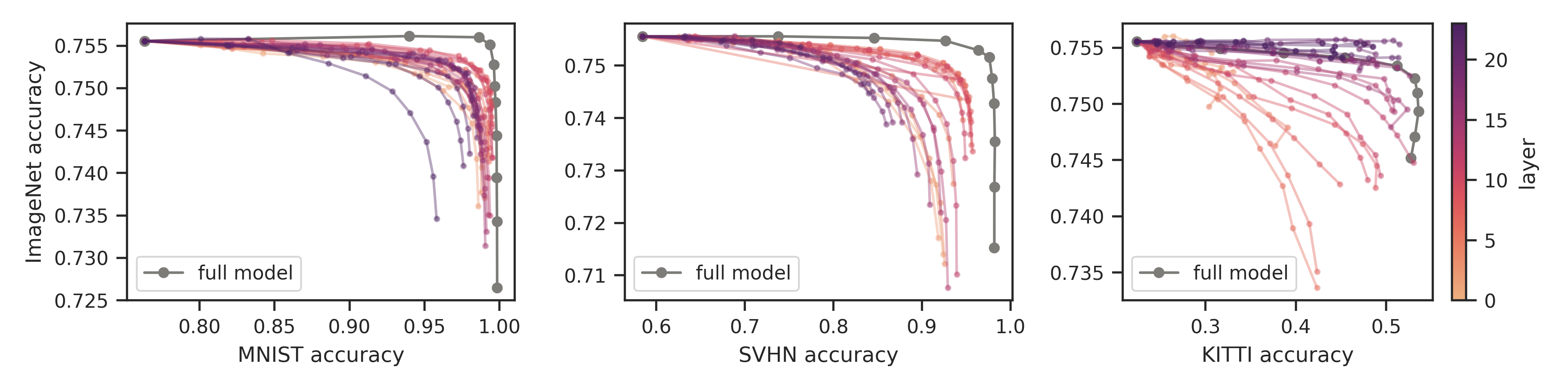}
  \caption{Linear interpolation between original and edited weights \(W \) and \(W'\) of CLIP models edited on various tasks. Each curve corresponds to a an editing layer (or to global fine-tuning), and each point on that curve corresponds to evaluation of a model with weights \((1-\alpha)W + \alpha W' \) for some \(\alpha \in [0, 1] \).}\label{fig:open-vocab-full-task}
\end{figure*}

\Cref{fig:open-vocab-full-task} shows a phenomenon similar to that of \cref{fig:single-layer-mli} occurs for a much larger CLIP open vocabulary model \cite{Radford2021LearningTV} on a variety of editing tasks: MNIST, KITTI and SVHN. When fine-tuning a single layer, ImageNet accuracy generally decreases as editing accuracy increases, and the resulting curves appear to be roughly concave. On the MNIST and SVHN tasks, the curve for global fine-tuning dominates those of single layer fine-tuning, but interestingly  we see that fine-tuning a single (later) layer can result in superior KITTI accuracy at a given ImageNet accuracy. Note also that for SVHN, tuning later layers results in worse performance than tuning earlier layers, indicating that the choice of editing layer is task dependent.

\begin{figure*}[bt]
   \centering
   \includegraphics[width=0.98\linewidth]{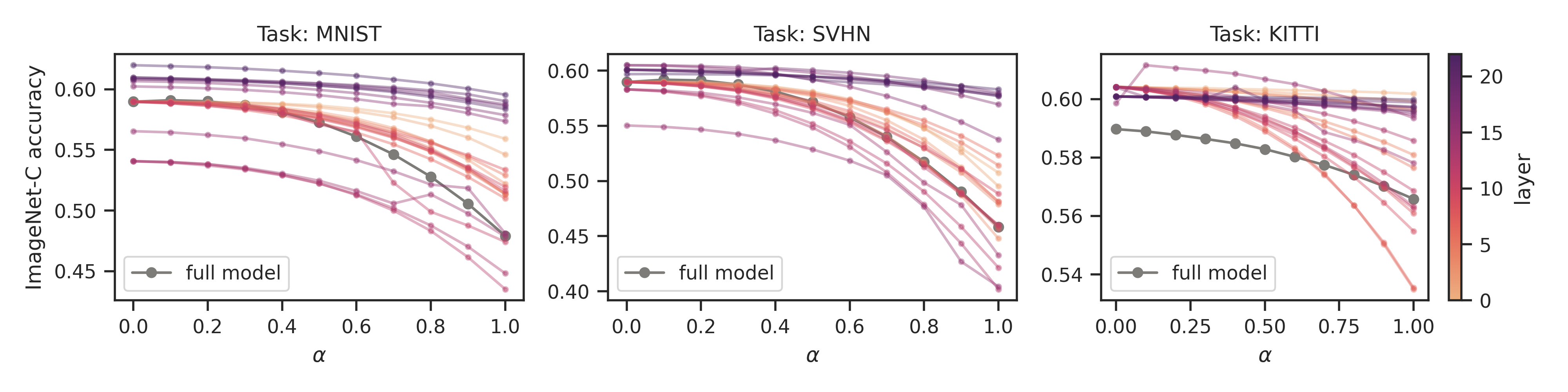}
   \caption{ImageNet-C accuracy along line segments \((1-\alpha) W + \alpha W' \) between original and edited weights \(W \) and \(W'\) of CLIP models edited on various tasks, plotted with respect to the interpolation parameter \(\alpha \).}\label{fig:1-L-imagenetC}
\end{figure*}

In addition to measuring accuracies on original and editing tasks, we can also measure \emph{OOD accuracy} along line segments \((1-\alpha)W + \alpha W'\). \Cref{fig:1-L-imagenetC} displays ImageNet-C accuracies of the CLIP models appearing in \cref{fig:open-vocab-full-task}, plotted with respect to the interpolation parameter \(\alpha\). Observe that these curves are generally monotonically \emph{decreasing}, showing that the edited models are less robust to corrupted variants of the original validation set than the original model. While in two of the three tasks considered, global fine-tuning dominated single-layer fine-tuning in terms of original and editing task accuracies (as seen in \cref{fig:open-vocab-full-task} (a, c)), we see that for all three tasks there are layers where single-layer fine-tuning provides higher ImageNet-C accuracy than global fine-tuning along some portion of the interpolation path. In the cases of MNIST and SVHN, this benefit only appears for \(\alpha \) near \(1\) (where interpolated weights are close to the edited weights), however for KITTI we see the effect at all measured values of \(\alpha \). \Cref{fig:1-L-imagenetR} displays results of a similar experiment, using ImageNet-R in place of ImageNet-C, from which we draw similar conclusions.

When we consider \cref{fig:INR-scatter}, as well as \cref{fig:open-vocab-full-task,fig:1-L-imagenetC,fig:1-L-imagenetR} as a whole, we see that while accuracy on OOD variants of ImageNet accuracy does appear to be linearly correlated with ImageNet accuracy, as is often observed to be the case \cite{rechtImageNetClassifiersGeneralize2019,taoriMeasuringRobustnessNatural2020,millerAccuracyLineStrong2021,wortsmanRobustFinetuningZeroshot2022}, there are notable exceptions. In particular, a simple linear correlation would predict global fine-tuning to exhibit higher ImageNet-C accuacy for all values of \(\alpha \) in \cref{fig:1-L-imagenetC} (a, b), which is evidently not the case. This deviation from linear correlation hints at a possibility that single-layer edited models may exhibit a form of \emph{effective robustness} as formalized in \cite{taoriMeasuringRobustnessNatural2020}. It also raises the question of whether that the finding that effective robustness decreases over the course of global or last-layer fine-tuning \cite{andreassenEvolutionOutofDistributionRobustness2021} holds in the case of local fine-tuning of early layers. Further investigation of these last few points would be an exciting direction for future work.

\subsection{New monotone linear interpolation phenomena}\label{sec:newmli}
One can view the results of \cref{fig:single-layer-mli,fig:open-vocab-full-task} as a new observation of a monotone linear interpolation (MLI) phenomenon. The original observations of MLI found empirically that training loss decreased monotonically along line segments between randomly initialized weights and weights at the endpoint of SGD training \cite{goodfellow2014qualitatively}. The basis for weight-space interpolation as a fine-tuning mechanism (as employed in \cite{ilharcoPatchingOpenvocabularyModels2022,wortsmanRobustFinetuningZeroshot2022}) is another form of MLI: empirically, as the interpolation parameter \(\alpha \) increases, accuracy on the fine-tuning task increases monotonically while accuracy on the original task decreases monotonically. \Cref{fig:single-layer-mli} shows the same behaviour occurs when weights are updated using local (i.e. single layer) fine-tuning or low-rank editing methods. We believe this finding is new, with the following noteworthy exception: one can prove that if only the final (pre-logit) weights are updated and the pairwise loss function is convex (for example, cross entropy loss) then MLI holds. We also argue that it was not \emph{a priori} obvious that single layer optimization loss landscapes share the same MLI-type behaviour observed for full model training and fine-tuning.

It is worth mentioning \cite{lucasAnalyzingMonotonicLinear2021a} exhibits realistic examples of neural network training where MLI fails, even though models train successfully --- a common theme of these examples is that they involved long-distance travel of model weights (for example, using the Adam optimizer \cite{kingmaAdamMethodStochastic2017} reliably broke MLI). We suspect that a reason failures of MLI are not more commonly observed in the context of fine-tuning and editing is the common practice of limiting the distance travelled by weights in these situations (see for example the discussion in \S 4.2, 8 of  \cite{ilharcoPatchingOpenvocabularyModels2022}).

\begin{figure*}[bt]
   \centering
   \includegraphics[width=0.98\linewidth]{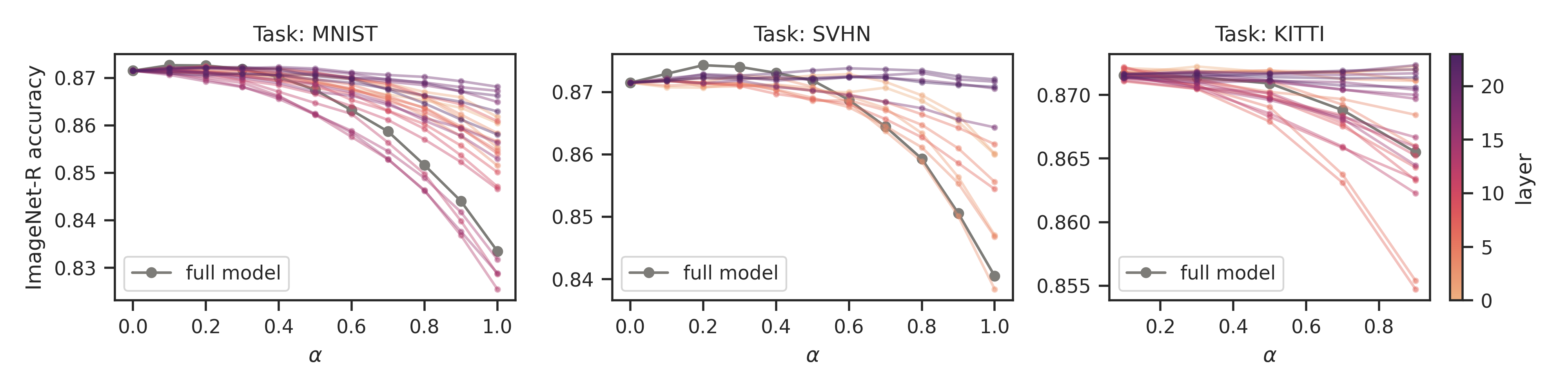}
   \caption{Same as \cref{fig:1-L-imagenetC}, but with ImageNet-R accuracy along line segments \((1-\alpha) W + \alpha W' \) between original and edited weights \(W \) and \(W'\) of CLIP models edited on various tasks, plotted with respect to the interpolation parameter \(\alpha \).}\label{fig:1-L-imagenetR}
\end{figure*}

\section{Limitations}



There are a range of important ways that the present study could have been made more comprehensive. Large language models are an important application area of editing techniques which we were not able to cover in this work. Likewise, there are other notions of distribution shift and methods for measuring distribution shift. 
Prior work on model editing has studied aspects of the generalization capabilites of edited models on inputs related to the editing task not considered in this work, where we chose to focus on robustness to shifts of natural image distribtions. In particular, it would be interesting to know whether or not direct low-rank editing and 1-LI enjoy the same  benefits as rewriting in the cases where the editing training and validation set are derived from ImageNet images with disjoint underlying sets of labels (see Figs. 3, 5 of \cite{santurkarEditingClassifierRewriting2021a}).
Finally, in performing our experiments, a large number of hyperparameters needed to be chosen. While we strove to optimize all of these, it remains possible that further refinement may have improved the performance of some of the editing methods.

\section{Conclusion}

In our experiments
there appears to be at least one common thread: editing weight updates with restricted capacity (as measured by the effective number of model parameters the update modifies) can preserve more of the robustness of the original model. Indeed, in \cref{fig:corr-lr-ft} we see that rewriting, a method that perturbs a single layer's weight matrix with a low-rank matrix, incurrs a smaller (i.e. less negative) original task OOD penalty than full model fine-tuning, which is free to modify all network parameters. In \cref{fig:1-L-imagenetC,fig:1-L-imagenetR} we see that 1-LI, a method that by definition only modifies weights at a single network layer, retains higher accuracy on ImageNet-C and ImageNet-R than full model fine-tuning for many editing layers. Of course, in the case of 1-LI there are also layers that fare worse on ImageNet-C and ImageNet-R than full model fine-tuning, illustrating that editing weight update capacity cannot be the only factor in play.

We hope that these results can be used to better inform the choice of editing method in real-world situations where a model is expected to encounter out-of-distribution data upon deployment.

\section{Acknowledgments}
This research was supported by the Mathematics for Artificial Reasoning in Science (MARS) initiative at Pacific Northwest National Laboratory.
It was conducted under the Laboratory Directed Research and Development (LDRD) Program at at Pacific Northwest National Laboratory (PNNL), a multiprogram
National Laboratory operated by Battelle Memorial Institute for the U.S. Department of Energy under Contract
DE-AC05-76RL01830.

The authors would also like to thank Xander Davies and Tony Chiang for useful discussions related to this work.

\printbibliography

\clearpage
\appendix

\section{Broader impacts}

Much of the motivation for editing algorithms stems from the scale of nascent foundation models, and the desire for inexpensive weight update methods for changing their behavior. There are a host of unanswered questions and valid concerns about how these models will impact society --- rather than enumerating a random sample here we refer to \cite{Bommasani2021FoundationModels} for a thorough discussion. As with any model update technique, editing algorithms have the potential for abuse, such as the insertion of backdoor triggers \cite{guBadNetsIdentifyingVulnerabilities2019}; our 1-LI and direct low-rank algorithms are not exceptions.

We are not aware of any potential negative consequences of our robustness evaluations of edited neural networks. On the contrary, as we write this there has been a rapid proliferation of deep learning models obtained by fine-tuning and editing a relatively small number of publicly available pretrained transformers on a plethora of tasks, and we hope our experiments shed light on overlooked side-effects of this paradigm. 

\section{Editing methods}
\label{sec:edmeth}

In this section we provide a more in-depth description of the editing methods described in \cref{sec:edit-methods}.  We find it conceptually useful to organize the discussion around two tools in model editing: methods for updating model weights and methods for balancing the trade off in accuracy between the original and editing task. Most of the references in \cref{sec:rw} apply some weight update procedure together with cross validation, however \cite{ilharcoPatchingOpenvocabularyModels2022} recently obtained impressive results using global fine-tuning and weight space linear interpolation. Further details, including notes on implementation, are deferred to \cref{sec:editing-method-details}. 

\textbf{Methods of updating weights:} Consider an input pair \((x, x')\), as in the vehicles on snow or large synthetic tasks (introduced in \cref{sec:tasks-data}).
If \(f_{\leq l}(x) = f_{\leq l}(x')\) for any \(l\), it trivially follows that \(f(x) = f(x')\), as desired. 
This is the basic motivation for the following weight update procedures, all of which attempt to minimize the mean squared error between \(f_{\leq l}(x)\) and \( f_{\leq l}(x')\) creating the situation where \(f_{\leq l}(x) \approx f_{\leq l}(x')\):
\begin{enumerate}[i)]\itemsep0em
    \item \label{item:loc-ft-oc} \emph{Local fine-tuning for output collision}: Optimize \(W_l\) using SGD.
    \item \emph{Global fine-tuning for output collision}: Optimize \(W_1, \dots, W_l\) using SGD.
    \item  \label{item:rewrit} \emph{Rewriting}
    \cite{bauRewritingDeepGenerative2020,santurkarEditingClassifierRewriting2021a,mengLocatingEditingFactual2022,mengMassEditingMemoryTransformer2022}: Perform update \(W_l \gets W_l + U V^T\), where \(U, V \) are low-rank matrices.
    \(V\) is derived by viewing \(W_l\) as a linear associative memory \cite{kohonenCorrelationMatrixMemories1972,andersonSimpleNeuralNetwork1972a,kohonenAssociativeMemory1977} and \(U\) is optimized using SGD.
\end{enumerate} 
For fine-tuning tasks where updates use pairs $(x',y)$ rather than $(x,x')$, additional methods are available:
\begin{enumerate}[i),resume]\itemsep0em
    \item \label{item:loc-ft} \emph{Local fine-tuning at layer $l$}: Optimize \(W_l\) with SGD such that \(f(x) \approx y\) for new datapoints \((x, y)\).
    \item  \label{item:glob-ft} \emph{Global fine-tuning from layer $l$ forward}: Optimize \(W_l, \dots, W_L\) with SGD such that \(f(x) \approx y\) for new datapoints \((x, y)\)
\end{enumerate} 

\textbf{Approaches to balancing original and editing accuracy}: The two dominant approaches that we will consider are: 
\begin{enumerate}[i)]\itemsep0em
    \item \emph{Cross validation}: In this simple approach, one monitors the validation accuracy on the original and editing validation datasets while performing optimization with respect to the editing task. One chooses a desirable ``best iteration'' (i.e. early stopping) as suits the application.
    \item \emph{Weight space linear interpolation}: Surprisingly, linearly interpolating weights often interpolates between performance on the original and editing tasks. Specifically, given original weights $W := (W_l)^L_{l = 1}$ and edited weights  $W' := (W_l')^L_{l = 1}$ one chooses $\alpha \in [0,1]$ such that $(1-\alpha)W + \alpha W'$ balances performance between the two tasks.
\end{enumerate}

We note that weight space interpolations have only been explored for full-model fine-tuning. To our knowledge we are the first to observe that linear interpolation works when weights have been modified using an editing method that only updates weights at a single layer (see Section \ref{sec:newmli}).

Finally, we introduce two new editing methods that we discovered during our experiments on editing robustness.

\textbf{Single-layer interpolation (1-LI)}: We apply local fine-tuning at layer \(l\) along with weight-space interpolation. The inspiration for this approach was to combine a common feature of several editing algorithms, namely weight updates restricted to a single layer, with the empirical benefits of weight-space interpolation \cite{ilharcoPatchingOpenvocabularyModels2022, wortsmanRobustFinetuningZeroshot2022}. In Section \ref{sec:1lirob} we show that this method often reduces degradation of model robustness. 

\textbf{Direct low-rank editing}: The motivation for our direct low-rank editing algorithm comes from asking how much of the effectiveness of other model editing methods stem \emph{simply from employing low-rank weight updates}, and our experimental results in Section \cref{sec:rob-degrade}\cref{sec:rob-depends} suggest that at least when comparing with rewriting, the answer is that much of the effectiveness is due to rank restriction alone. 
We provide additional details, contrast the method with those of 
\cite{bauRewritingDeepGenerative2020,santurkarEditingClassifierRewriting2021a}
and provide a code sample in \cref{sec:editing-method-details}. 

\section{Direct low-rank editing implementation}
\label{sec:editing-method-details}

In this section we only provide a complete description of direct low-rank editing. The rewriting algorithm is described in detail in \cite{bauRewritingDeepGenerative2020} (see also \cite{santurkarEditingClassifierRewriting2021a}, which adapts rewriting to image classifiers). We take fine-tuning to be well known (or easy to locate in existing literature), and emphasize 1-LI differs from the PAINT method of \cite{ilharcoPatchingOpenvocabularyModels2022} only in the restriction of weight updates during fine-tuning to a single layer.  

Direct low-rank editing is related to two successful editing methods, rewriting and the MEND method \cite{mitchellFastModelEditing2022}, that employ low-rank perturbations to weight matrices. Beyond this common feature of low-rank perturbations, the two methods differ in many other details: MEND trains hypernetworks to optimize its low-rank perturbations \(U V^T \) and initializes \(U V^T \) by applying a singular value decomposition to model gradients on batches of editing samples\footnote{Which are naturally low-rank, in fact with rank bounded by the batch size (see e.g. Appendix E of \cite{kianiProjUNNEfficientMethod2022})} --- this stands in contrast to the description of rewriting appearing in \cref{item:rewrit} above.\footnote{We do not compare with MEND in this work since it has somewhat different data requirements than those described in \cref{sec:tasks-data} --- in particular, an auxiliary dataset of editing examples is needed to train the hypernetwork to edit models. Evaluating direct low-rank editing on the tasks considered in \cite{mitchellFastModelEditing2022} would be an interesting experiment for future work.}

In the case of direct low-rank editing at layer \(l \) of a network \(f\) as in \cref{eq:toy-nn}, we introduce new parameters \(U, V \) where \(U \) (resp. \(V\)) is an \(n_{l} \times r \) (resp. \(r \times n_{l-1} \)) matrix; here \(r \) is the rank of the edit, a hyper parameter decided in advance. Then, with all weights \(W_1, \dots, W_L \) frozen, we optimize the network weights 
\begin{equation}
    (W_1, \dots, W_{l-1}, W_l + UV^T, W_{l+1}, \dots, W_L )
\end{equation}
to minimize the mean squared error \( \frac{1}{n_l} \lvert f_{\leq l}(x) - f_{\leq l}(x') \rvert^2\) on the feature collision style editing training set using SGD on the parameters \(U, V \).\footnote{Note that since no earlier layers are trained, this is equivalent to minimizing the mean squared error between \(\sigma((W_l + UV^T) f_{\leq l-1}(x)) \) and \(\sigma((W_l + UV^T) f_{\leq l-1}(x')) \), where the \(l-1\)st features \(f_{\leq l-1}(x)\) and \(f_{\leq l-1}(x')\) are fixed throughout editing optimization.} A few notes:
\begin{itemize}
    \item This method shares the low-rank feature of rewriting.
    \item It lacks a principled derivation through the lens of associative memories. 
    \item It is potentially computationally less expensive: for example, rewriting requires pre-computation of the covariance matrix \(\Sigma \) of the features \(f_{\leq l-1}(x) \) over datapoints in the original training set \(\cD_{\text{orig}}^{\text{train}} \) and computing its inverse square root (to later apply a ZCA transformation). In the case of today's large neural networks, this feature covariance matrix can be quite large (e.g. \(4096 \times 4096\)). Direct low-rank editing requires no preliminary numerical linear algebra. However, it is also true that a careful implementation of rewriting can avoid a large matrix inversion by the classic trick of replacing a ``matrix-inverse-vector'' operation \(A^{-1} y \) with a call to a least squares solver for \( \nrm{Ax - y}^2 \), and we have not formally benchmarked the computational cost of direct low-rank and rewriting. 
\end{itemize}


Most of the image classifiers we experiment with are CNNs, and in this case with the exception of the last few layers we are editing 2D \emph{convolution} weights, rather than the matrices appearing in our toy model \cref{eq:toy-nn}. All editing methods other than direct low-rank and rewriting extend essentially verbatim to the case of CNNs. For direct low-rank and rewriting, we implement the low-rank perturbations as \(1\times1\) convolutions (equivalent to multiplying channel vectors with the same low-rank matrix at every spatial position). More precisely, let \( W_l \) be the original, unedited convolution weight, say of shape \(C_l \times C_{l-1}\times K \times K \) where \(K \) is the kernel size. We then let \(U \) and \(V \) be convolutions of shapes  \(C_l \times r \times 1 \times 1\) and \(C_{l-1} \times r \times 1 \times 1\) respectively and replace the expression \(W_l + UV^T \) used for fully connected linear layers with 
\begin{equation}
\label{eq:conv-version}
    W_l + U \ast V^T
\end{equation}
where \(\ast \) denotes convolution and transpose flips the first two indices. Note that as \(1\times1 \) convolutions \(U \) and \(V^T \) simply apply a fixed matrix over the channels of an input hidden feature at each spatial position. In \cite{santurkarEditingClassifierRewriting2021a} it was noted that this improves performance of rewriting, when compared to applying the perturbation \(W_l + U \ast V^T\) only at spatial positions contained in the segmentation mask of the editing task (``road'' in the case of vehicles on snow, or the concept being replaced via style transfer in the case of the large scale synthetic task). 

A convolution using the weights of \cref{eq:conv-version} can be implemented in a few lines of idomatic PyTorch:

\begin{lstlisting}
import torch
from torch.nn import functional as F


# u, v are matrices of shape (C_l, r), (C_{l-1}, r) respectively
# multiply to get a rank <= r matrix:
uv = torch.einsum('i,j->ij', u,  v)
# reshape to get a 1x1 convolution weight:
uv = uv.reshape(uv.shape + (1, 1))
# w is the original convolution weight of shape (C_l, C_{l-1}, K, K)
# x is an input of shape (C_{l-1}, H, W)
output = F.conv2d(x, w) + F.conv2d(x, uv)
\end{lstlisting}

\Cref{eq:conv-version} only strictly-speaking makes sense when the kernel size \(K \) is \emph{odd}: in that case, we can pad the \(1\times1 \) convolutions \(U \) and \(V \) with zeros in the spatial tensor dimensions to obtain \(K\times K \) convolutions, say \(\tilde{U} \) and \(\tilde{V} \) such that if \((k,l) \) denotes the index at the center of the \(K\times K \) kernel (which depends on indexing conventions)
\begin{equation}
    \tilde{U}_{cc'ij} = \begin{cases}
    U_{cc'ij} & \text{ if } i=k, j=l \\
    0 & \text{ otherwise}
    \end{cases}
\end{equation}
and similarly for \(\tilde{V}\).
Then we can sensibly evaluate \(W_l + \tilde{U} \ast \tilde{V}^T \). In the case where \(K \) is even, if one wants an update that truly only modifies the weights of a network (rather than appending an auxiliary convolutional weight), there exist reasonable workarounds, for example if \(\{k, k+1\} \times \{l, l+1\} \) is the center of the \(2\times2\) grid at the center of the \(K\times K \) kernel (which again depends on indexing conventions) one could use the weight sharing scheme
\begin{equation}
    \tilde{U}_{cc'ij} = \begin{cases}
    U_{cc'ij} & \text{ if } i -k \leq 1, j-l \leq 1 \\
    0 & \text{ otherwise.}
    \end{cases}
\end{equation}
Thankfully, since the VGG and ResNet CNNs in our experiments only use odd kernel sizes we never have to resort to such measures. 

Lastly, following the discussion in \S A.4 of \cite{santurkarEditingClassifierRewriting2021a} for ResNets we only edit convolutional layers appearing at the end of residual blocks. 


\section{Experimental details}

\subsection{Synthetic editing datasets and corruptions thereof}
\label{sec:corr-lss}

We start with the segmented ImageNet validation datapoints\footnote{These segmentations were obtained with a DeepLabV2 segmentation model \cite{chenDeepLabSemanticImage2017} trained on the COCO-Stuff dataset \cite{caesarCOCOStuffThingStuff2018}.} and style images made available in the code release of \cite{santurkarEditingClassifierRewriting2021a} --- these segmented ImageNet validation datapoints are organized according to ``concepts'', i.e. segmentation labels assigned to a large region of the image (for example ``mountain''), and the style images are organized into groups (for exampe ``fall colors'') each of which contains several images. Then, for each concept-style pair \((c, s\) we generate a dataset of triples \((x, x', y) \) where \((x, y) \) is an ImageNet validation datapoint for which the concept occupies a large region in the segmentation map and \(x'\) is obtained by replacing the region of the image \(x\) segmented as the concept \(c\) with the style \(s\), using Magenta's ``arbitrary image stylization'' model \cite{ghiasiExploringStructureRealtime2017}. Images (a) and (b) of \cref{fig:corr-lss-ex} show an example datapoint. 
\begin{figure}
   \centering
   \begin{subfigure}{0.30\linewidth}
      \centering
      \includegraphics[width=\linewidth]{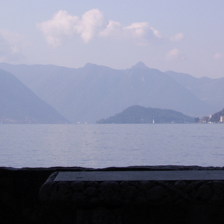}
      \caption{}
   \end{subfigure}
   \begin{subfigure}{0.30\linewidth}
      \centering
      \includegraphics[width=\linewidth]{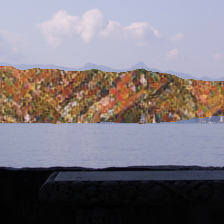}
      \caption{}
   \end{subfigure}
   \begin{subfigure}{0.30\linewidth}
      \centering
      \includegraphics[width=\linewidth]{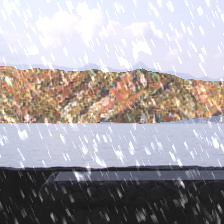}
      \caption{}
   \end{subfigure}
   \caption{An example of an editing datapoint created using segmentation and style transfer. Image (a) is an ImageNet validation datapoint, with label 976 (promontory, headland, head, foreland). In image (b), the concept ``mountain'' has been replaced with the style ``fall trees.'' In image (c), the corruption ``snow'' has been applied at a low severity level. Best viewed in color.}\label{fig:corr-lss-ex}
\end{figure}

The result of this pipeline is a large number of synthetic editing datasets \( \cD_{\text{edit}}(c,s) \): from 17 concepts and 8 styles we obtain 136 editing datasets, one per concept-style pair. The concepts and styles used are listed below.

\begin{description}
\item[Concepts] ``waterother'', ``clouds'', ``wallconcrete'', ``boat'', ``table'', ``mountain'', ``skyother'', ``diningtable'', ``light'', ``plantother'', ``tree'', ``sea'', ``person'', ``snow'', ``tv'', ``road'', ``grass''.
\item[Styles] ``falltrees'', ``snowyground'', ``woodentexture'', ``furrytexture'', ``blackandwhite'', ``colorfulflowers'', ``graffiti'', ``gravel''.
\end{description}

Our methodology for creating train-validation splits of such a synthetic dataset with concept \(c\) and style \(s\) is as follows. First, a number \(N_{text{train}} \) is specified (in all of our editing experiments on ImageNet models it is 10), along with a ``minimum training ratio'' \(\rho\) (in all our experiments this is \(1/2\)). In addition, a number \(S_{\text{train}} \) is specified and the different images associated with the style \(s \) are broken into train and validation splits (in all our experiments, \(S_{\text{train}} =1\), i.e. we include only one variant of the style \(s\) in the train split, and the rest in the validation split).  Then, we restrict attention to ImageNet class labels \(y \) with at least \(N_{text{train}} /\rho \) associated images in the full synthetic dataset with concept \(c\) and style \(s\). For each label \(y \), we randomly sample\footnote{Random sampling (including the random sampling of \(S_{\text{train}} \) training style variants) is controlled by a fixed random seed.} \(N_{\text{train}}\) corresponding datapoints \((x, y) \), and for each of those datapoints we add the \(S_{\text{train}} \) triples \( (x, x', y)\) where \(x' \) is obtained by replacing concept \(c \) with a training split variant of the style \(s \) to \(\cD_{\text{edit}}^{\text{train}}(c,s) \). The remaining triples \( (x, x', y)\) are added to \(\cD_{\text{edit}}^{\text{val}}(c,s) \). A few comments on this splitting strategy:
\begin{itemize}
    \item With our choices of the \(N_{text{train}} =10\), \(\rho=1/2\) and  \(S_{\text{train}} =1\) the resulting training set size is \(\nrm{\cD_{\text{edit}}^{\text{train}}(c,s)} = 10 \cdot\) the number of ImageNet labels with \(\geq 20 \) images in the dataset of segmentations provided by \cite{santurkarEditingClassifierRewriting2021a}. This means that the training data size varies with the concept \(c\) (but not with the style \(s \)). It also means that for every datapoint appearing in \(\cD_{\text{edit}}^{\text{val}}(c,s) \) there is a datapoint of \(\cD_{\text{edit}}^{\text{train}}(c,s) \) with the same label. Hence this editing task is in some sense easier than that of \cite{santurkarEditingClassifierRewriting2021a}, which required \(\cD_{\text{edit}}^{\text{train}}(c,s) \) to contain images from a single ImageNet class.
    \item \cite{santurkarEditingClassifierRewriting2021a} correctly points out that for some class labels \(y \), replacing the concept \(c \) with the style \(s \) makes classification (even by a human) difficult if not impossible. They manually filter such label-concept-style combinations from their analysis. We did not perform such a filtration. On the other hand, as mentioned above we do not restrict our editing training sets to come from a single ImageNet class, and as such it is unlikely that problematic label-concept-style combinations dominate our editing training sets. 
    \item In light of the above two points, note that we do not perform any analyses of how editing performance varies with editing training set size, nor any analyses of edited model performance on specific ImageNet classes.
\end{itemize}

We add corruptions to the validation sets \(\cD_{\text{edit}}^{\text{val}}(c,s) \) with the scripts used to generate ImageNet-C \cite{hendrycksBenchmarkingNeuralNetwork2019}. We use all 19 corruption types (including those denoted ``extra'' in ibid.) and all 5 severity levels. This results in 95 corrupted variants of each validation set \(\cD_{\text{edit}}^{\text{val}}(c,s) \). Image (c) in \cref{fig:corr-lss-ex} is an example of a corrupted concept-style-swapped image.

It must be acknowledged that in our evaluations of models edited using these synthetic datasets on ImageNet-C, there is a form of train-test leakage: the editing training sets \(\cD_{\text{edit}}^{\text{train}}(c,s) \) contain images from the (uncorrupted) ImageNet validation set, and our test set ImageNet-C consists of corrupted versions of the ImageNet validation set. This leakage is undesirable to say the least. In follow-on work, we would like to re-run these experiments and instead produce \(\cD_{\text{edit}}^{\text{train}}(c,s) \) from images in the ImageNet training set, or alternatively (if it is important to model the deployment scenario where new datapoints outside the original training set are collected) from images in ImageNetV2. Note in the later case we would face leakage evaluating edited models on ImageNetV2.

We note, however, that the train-test leakage in our experiments was limited: in the worst case (COCO-Stuff label ``person''), an edited model was exposed to \(1.04\% \) of the ImageNet validation set, and in the average case only \(0.215 \% \). Moreover, there are no literally identical images in the editing training set and ImageNet-C (although there certainly are corrupted versions of training images in ImageNet-C). 

\subsection{Editing method hyperparameters}

All experiments in this paper are implemented in PyTorch \cite{pytorch} (the one exception being the Magenta neural style transfer for synthetic editing dataset creation, where we use the original TensorFlow \cite{tensorflow2015-whitepaper} implementation) and run on NVIDIA GPUs. 

In the editing methods of \cref{sec:edit-methods}, wherever applicable we use PyTorch's \texttt{torch.optim.SGD} with momentum \(0.9\) and decay \(10^{-4}\) (only applied to parameters whose names contain ``weight''). We do not update batch norm running means and variances during editing. Our reasoning is that over the course of editing optimization those statistics will move from those of the original training dataset towards those of the editing dataset, which could have a destructive on model performance on the original task or corrupted variants thereof. We allow editing optimization to proceed for a maximum of 10000 epochs, with an early stopping criterion that terminates optimization if the editing validation accuracy drops below half of its best-so-far. The large number of maximum epochs is chosen to account for the huge variation in optimal learning rates observed at different layers (or sets of layers) of different models; the idea here is to gaurd against the possibility of never observing editing learning do to using a too-small learning rate (which did occur in many preliminary test runs). The early stopping criterion is in place to offset the high worst-case computational cost of using 10000 epochs in every single edit --- it seems to detect editing overfitting (which occurs quite regularly on the small datasets involved) in many cases, terminating optimization well before the maximum possible number of epochs is attained.

As mentioned above, effective\footnote{Here we say ``effective'' since optimal would be too strong (we do not do any fine-grained cross validation). By effective learning rate we just mean a learning rate that works, as in it decreases training loss (and hopefully increases validation accuracy in the process).} learning rates for the editing methods considered appear to span many orders of magnitude (based on preliminary tests, as low as \(10^{-5}\) in some cases of supervised fine-tuning of later layers and as high as 100 in some cases of direct low-rank and rewriting). Effective learning rates depend on model, editing dataset, editing method and layer (for the editing methods where a model layer is specified). We first use preliminary tests on a smaller set of editing datasets including vehicles on snow and synthetic datasets for a subset of concept-style pairs, and where relevant with a subset of possible editing layers, to determine for each model and editing method a reasonable interval in which to search for learning rates. Then in our experiments we choose learning rates using cross validation over a grid search in the relevant reasonable learning rate interval. 

In the case direct low-rank and rewriting, editing validation accuracy can exhibit substantial noise with respect to random initialization (of \(U\) and \( V \) in the case of direct low-rank, and of \(U \) in the case of rewriting). For these methods we run editing optimization several (\(\approx 10 \)) times with different random initializations (within the learning rate hyperparameter search).

We record the best observed editing validation accuracy and save the corresponding best set of edited model weights for downstream robustness evaluations.

Our implementation of rewriting is a ``from scratch'' rewrite: here we use quotes since we of course heavily reference \cite{bauRewritingDeepGenerative2020,santurkarEditingClassifierRewriting2021a} and the \texttt{EditingClassifiers} codebase available at \href{https://github.com/MadryLab/EditingClassifiers}{https://github.com/MadryLab/EditingClassifiers}. The choice to use a custom implementation was based on multiple factors, including the goal of porting rewriting to image classifiers not appearing in \texttt{EditingClassifiers} (namely Vision Transformers), and a desire to have a relatively uniform functional interface for the various editing methods in our experiments. We plan to open source our research code.

Unfortunately, there was a discrepancy between our version of rewriting and that in \cite{bauRewritingDeepGenerative2020,santurkarEditingClassifierRewriting2021a}: specifically, the code used in our experiments applied mean centering the the features denoted \(K \) in \cite[\S 3.3, eq. 14-15]{bauRewritingDeepGenerative2020}, thus essentially forcing \(KK^T\) to be a covariance matrix, rather than the second moment matrix as stated in ibid.\footnote{One source of confusion was the later derivation of editing feature rank reduction in \cite[\S D]{bauRewritingDeepGenerative2020} which idealizes the hidden features of the original (non-editing) dataset as a mean-zero Gaussian distribution, although it is never suggested to force them to be mean-zero in an implementation.} The code at \href{https://github.com/davidbau/rewriting}{https://github.com/davidbau/rewriting} and \href{https://github.com/MadryLab/EditingClassifiers}{https://github.com/MadryLab/EditingClassifiers} clearly does not apply mean-centering. We are in the process of assessing the sensitivity of the results for the rewriting method presented above to this mean centering issue. One reason to suspect an application of mean centering does not make a large difference is that all layers of all models edited in our experiments are either immediately preceded by a batch normalization (or in the case of ViTs, a layer normalization) layer, or have a normalization layer within the preceding 3 or so layers. The nearby presence of normalization layers would potentially limit the magnitude of feature means. 

For our PAINT \cite{ilharcoPatchingOpenvocabularyModels2022} and our 1-LI interpolation and runs, we adapt the code and training hyperparameter defaults from \href{https://github.com/mlfoundations/patching}{https://github.com/mlfoundations/patching} unless stated otherwise.



\section{Addition single layer interpolation (1-LI)}
The performances of the full-model interpolations and single-layer interpolations (1-LI) for individual ImageNet-C corruptions are provided here. Figure \ref{fig:1-L-imagenetC-breakdown} provides the individual corruption performances vs interpolation parameter $\alpha$, which is aggregated in Figure \ref{fig:1-L-imagenetC}. Similarly, Figure \ref{fig:1-L-imagenetC-vs-task-breakdown} plots the corruption performances against the task performances along the interpolation parameters.

\begin{figure}[b]\ContinuedFloat
\begin{center}
\begin{subfigure}[b]{0.95\textwidth}
   \includegraphics[width=1\linewidth]{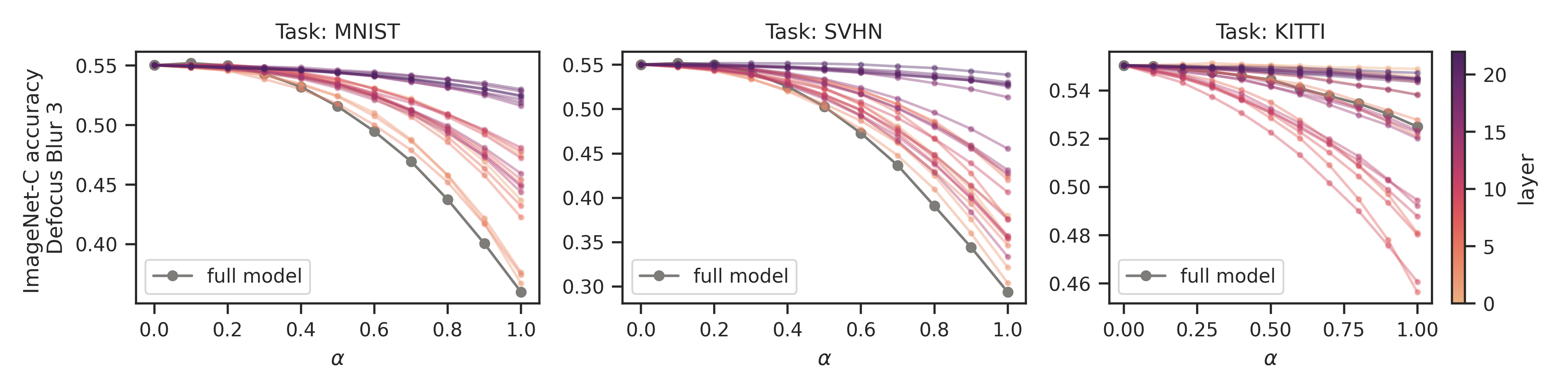}
\end{subfigure}
\begin{subfigure}[b]{0.95\textwidth}
   \includegraphics[width=1\linewidth]{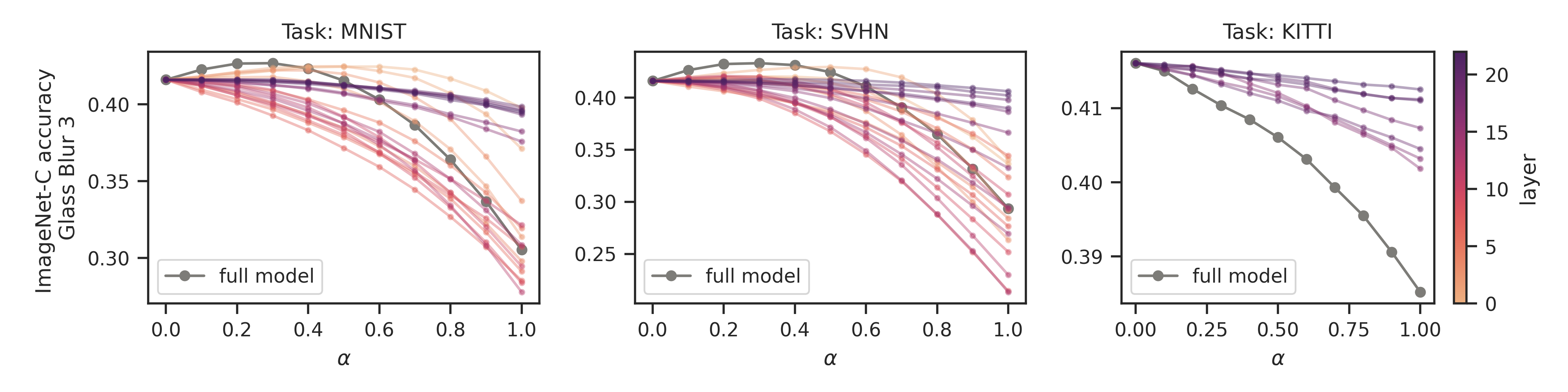}
  \label{fig:1-L-imagenetR}
\end{subfigure}
   \caption{(continued below)}\label{fig:1-L-imagenetC-breakdown}
 \end{center}
\end{figure}

\begin{figure}\ContinuedFloat
\begin{center}
\begin{subfigure}[b]{0.95\textwidth}
   \includegraphics[width=1\linewidth]{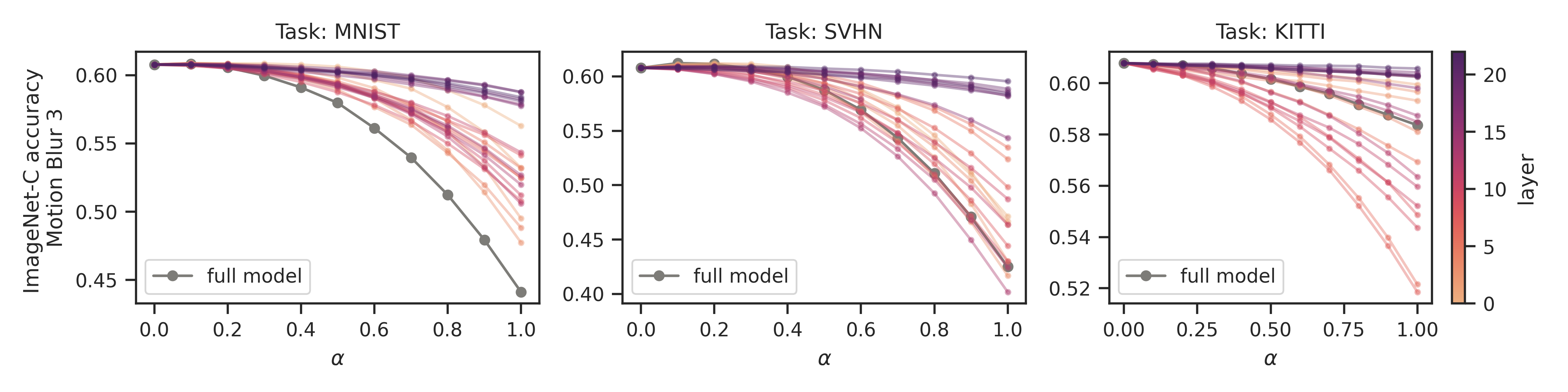}
  \label{fig:1-L-imagenetR}
\end{subfigure}
\begin{subfigure}[b]{0.95\textwidth}
   \includegraphics[width=1\linewidth]{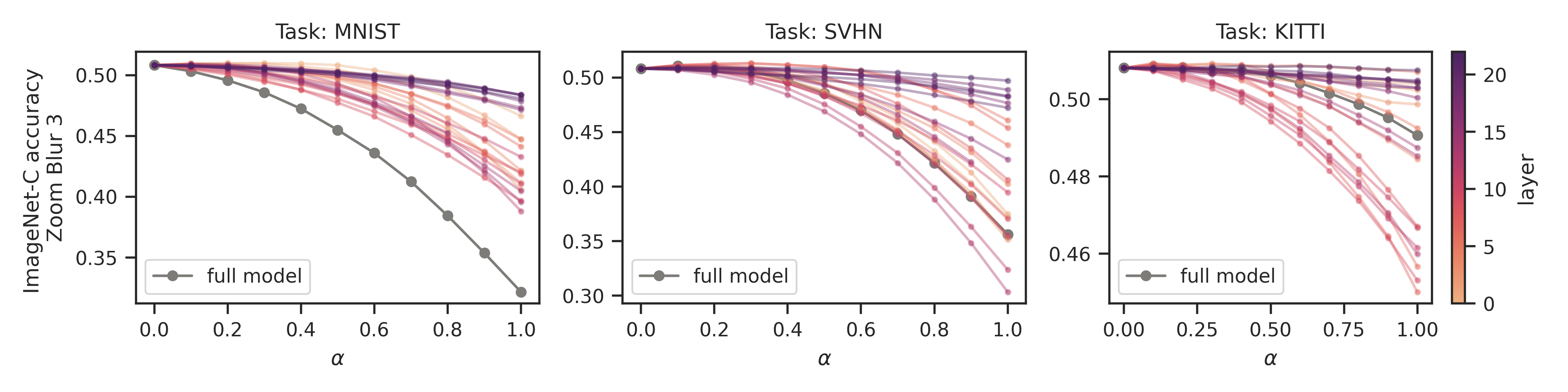}
  \label{fig:1-L-imagenetR}
\end{subfigure}
\begin{subfigure}[b]{0.95\textwidth}
   \includegraphics[width=1\linewidth]{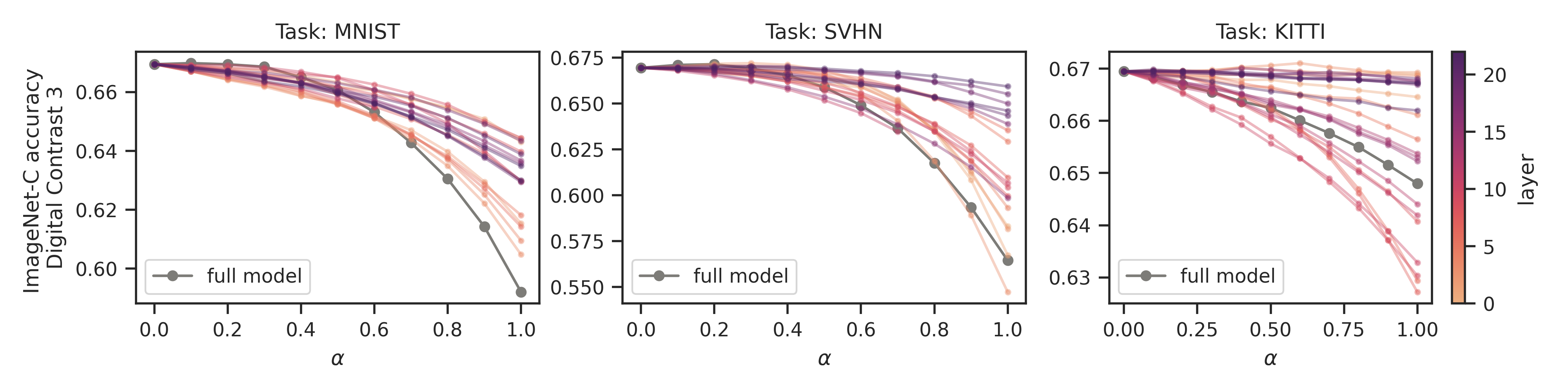}
  \label{fig:1-L-imagenetR}
\end{subfigure}
\begin{subfigure}[b]{0.95\textwidth}
   \includegraphics[width=1\linewidth]{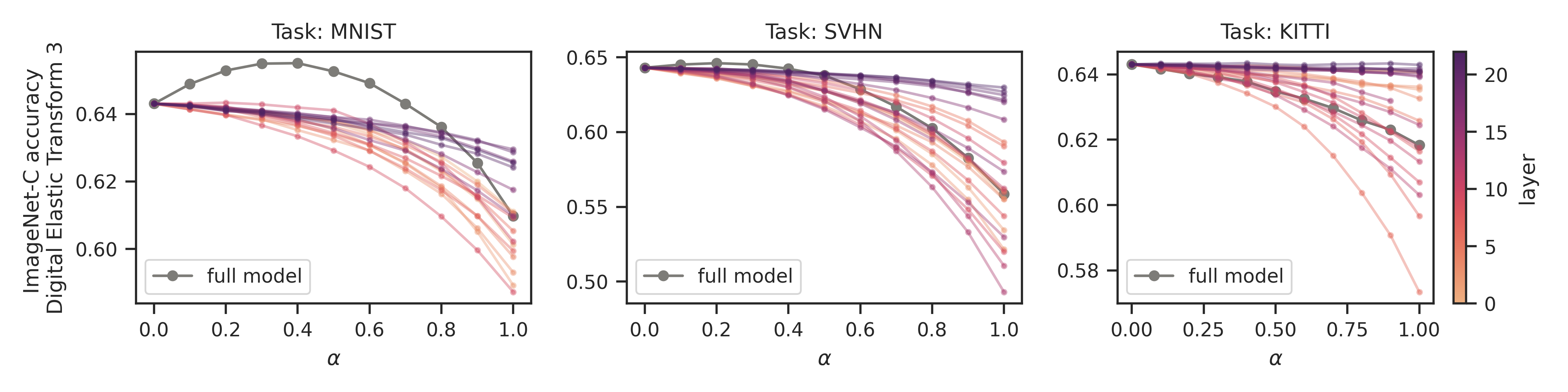}
  \label{fig:1-L-imagenetR}
\end{subfigure}
\begin{subfigure}[b]{0.95\textwidth}
   \includegraphics[width=1\linewidth]{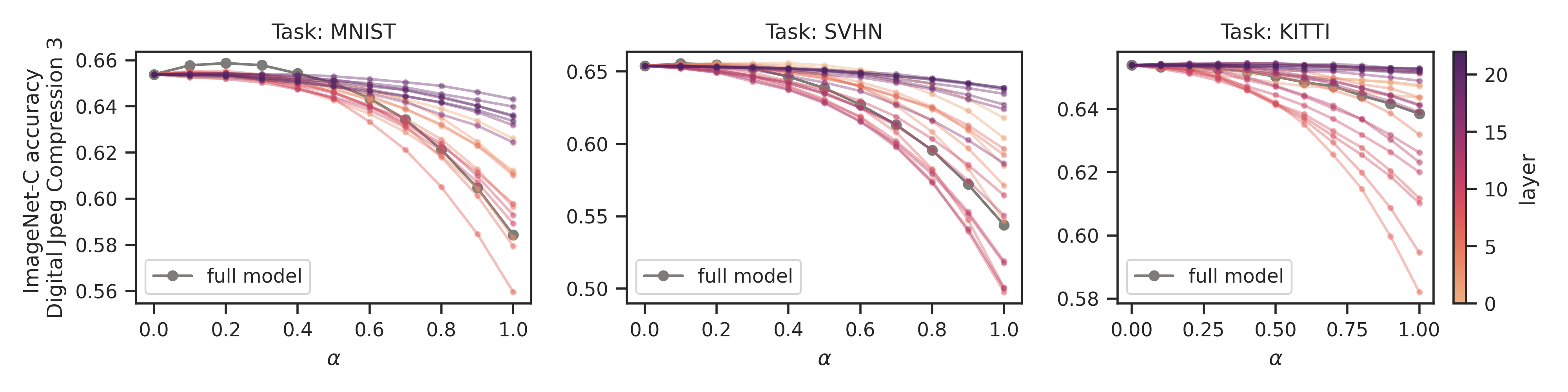}
  \label{fig:1-L-imagenetR}
\end{subfigure}
\begin{subfigure}[b]{0.95\textwidth}
   \includegraphics[width=1\linewidth]{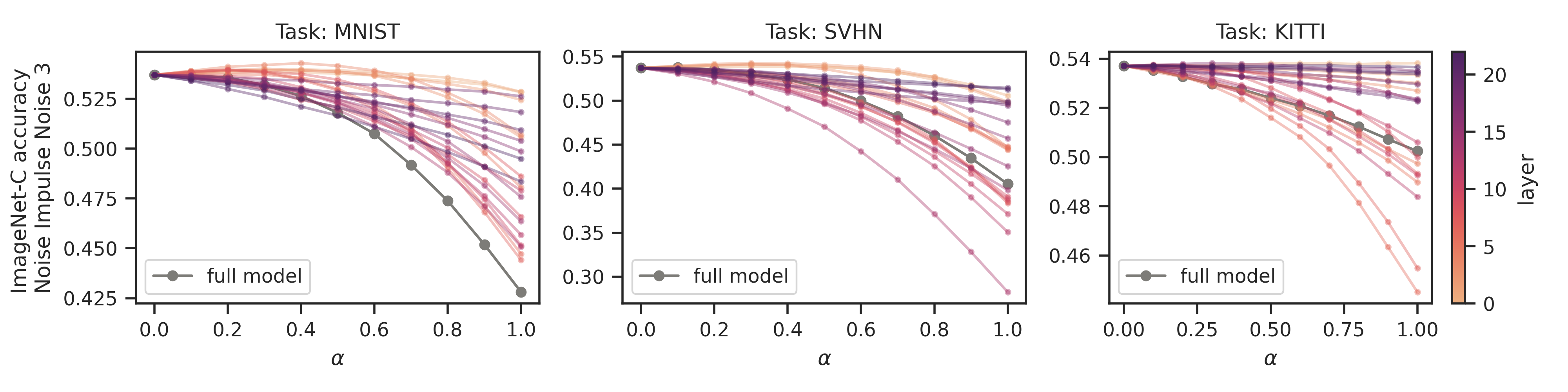}
  \label{fig:1-L-imagenetR}
\end{subfigure}
   \caption{(continued below)}\label{fig:1-L-imagenetC-breakdown}
 \end{center}
\end{figure}

\begin{figure}\ContinuedFloat
\begin{center}

\begin{subfigure}[b]{0.95\textwidth}
   \includegraphics[width=1\linewidth]{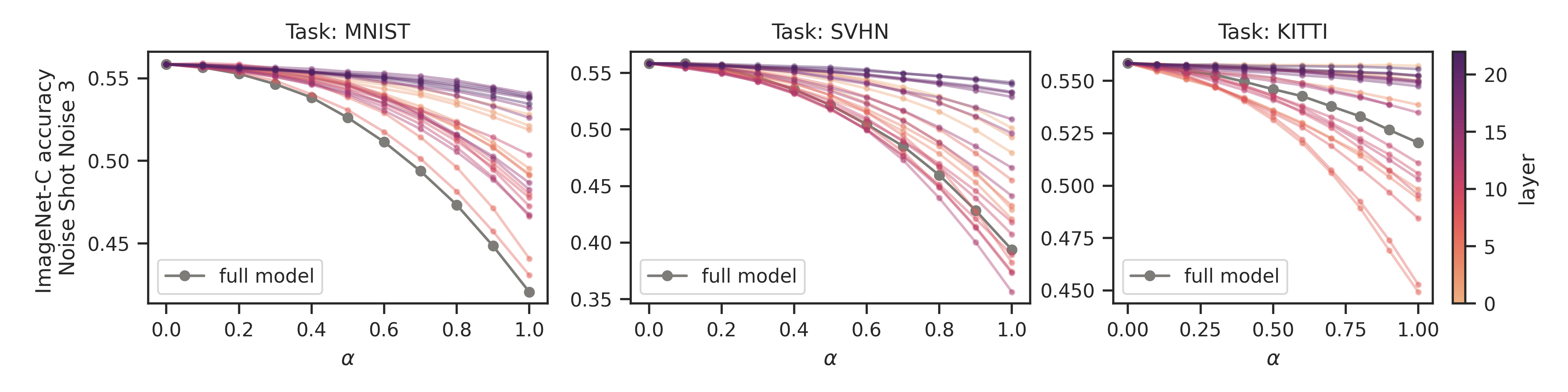}
  \label{fig:1-L-imagenetR}
\end{subfigure}
\begin{subfigure}[b]{0.95\textwidth}
   \includegraphics[width=1\linewidth]{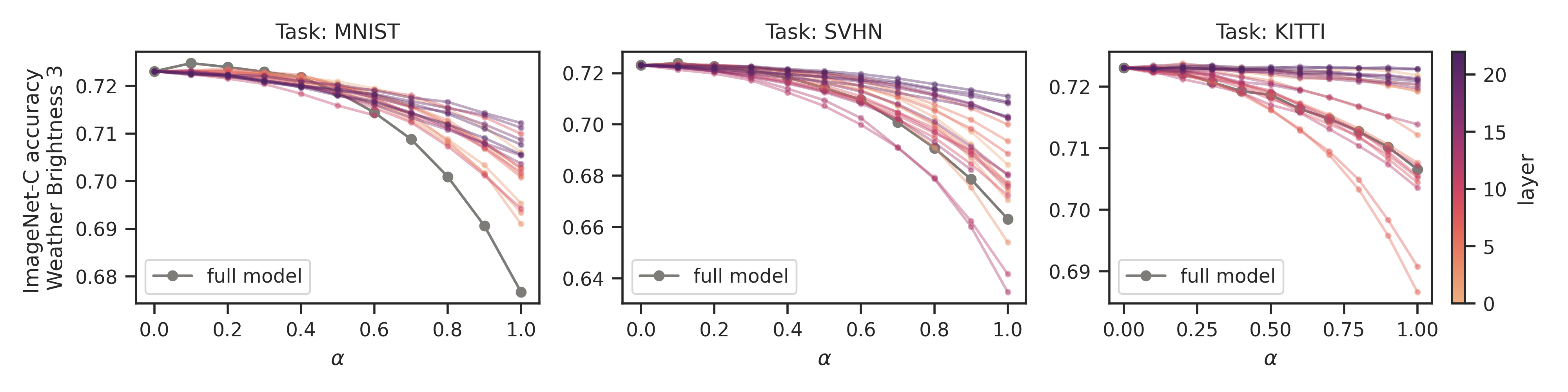}
  \label{fig:1-L-imagenetR}
\end{subfigure}
\begin{subfigure}[b]{0.95\textwidth}
   \includegraphics[width=1\linewidth]{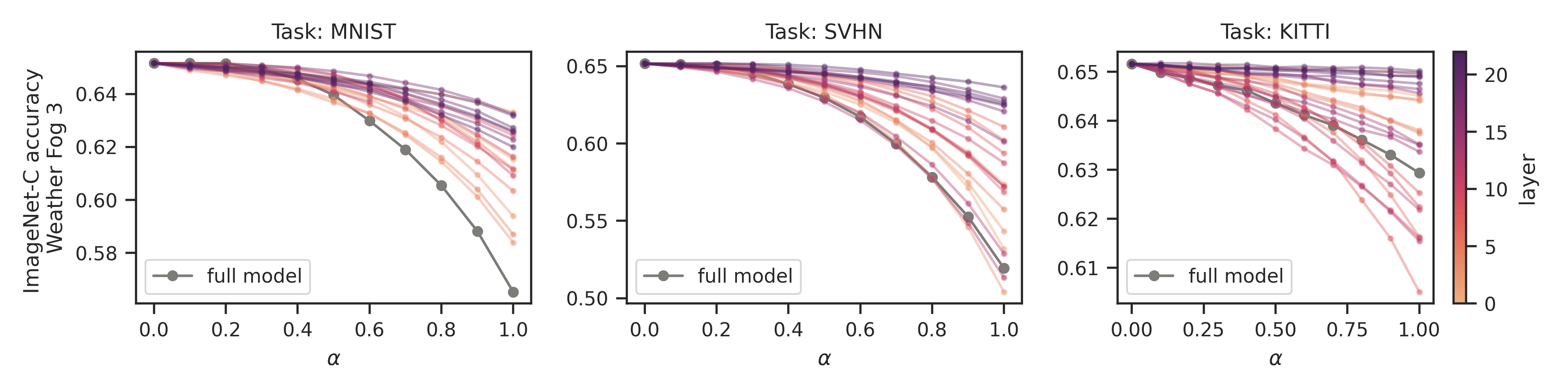}
  \label{fig:1-L-imagenetR}
\end{subfigure}
\begin{subfigure}[b]{0.95\textwidth}
   \includegraphics[width=1\linewidth]{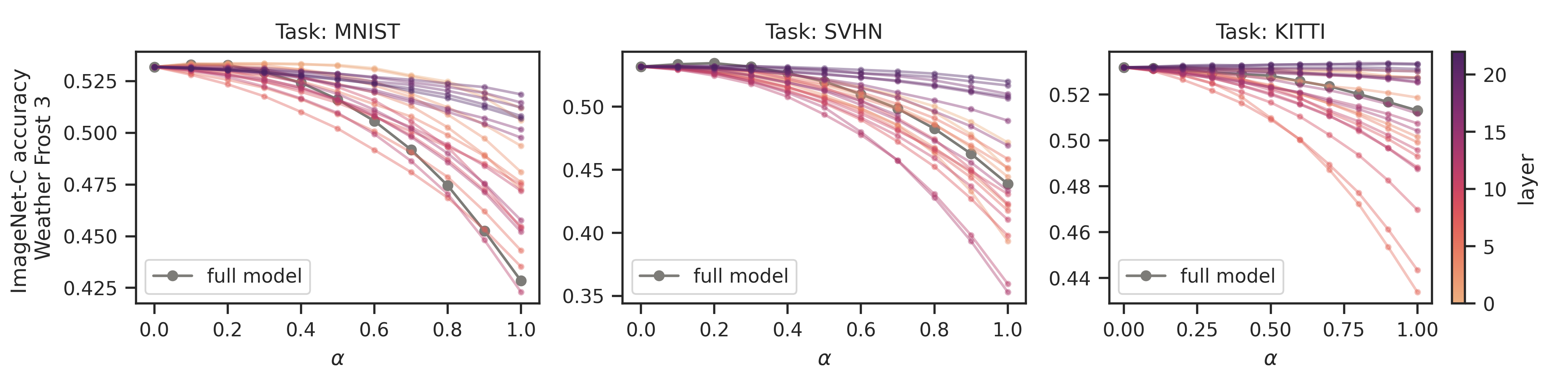}
  \label{fig:1-L-imagenetR}
\end{subfigure}
\begin{subfigure}[b]{0.95\textwidth}
   \includegraphics[width=1\linewidth]{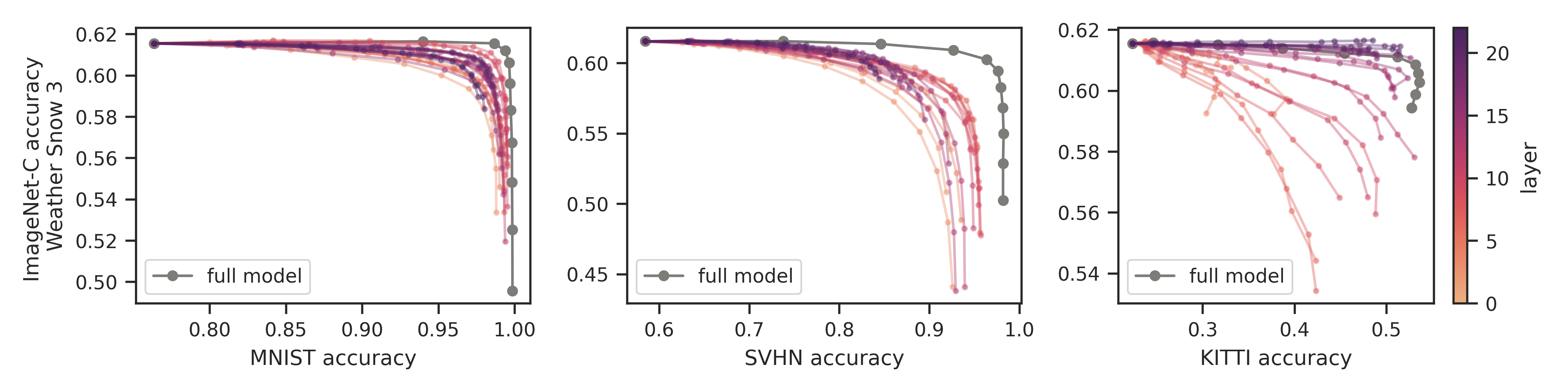}
  \label{fig:1-L-imagenetR}
\end{subfigure}
   \caption{Specific ImageNet-C corruption accuracies, for a corruption strength of 3, aggregated by Fig. \ref{fig:1-L-imagenetC}. Computed along line segments \((1-\alpha) W + \alpha W' \) between original and edited weights \(W \) and \(W'\) of CLIP models edited on various tasks, plotted with respect to the interpolation parameter \(\alpha \).}\label{fig:1-L-imagenetC-breakdown}
 \end{center}
\end{figure}

\begin{figure}[b]
\begin{center}
\begin{subfigure}[b]{0.95\textwidth}
   \includegraphics[width=1\linewidth]{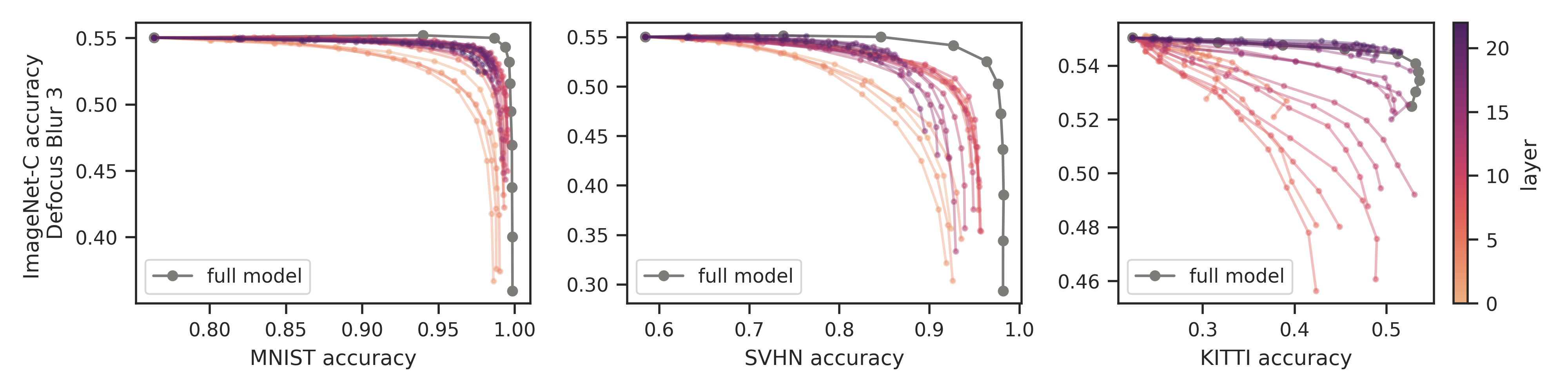}
\end{subfigure}
\begin{subfigure}[b]{0.95\textwidth}
   \includegraphics[width=1\linewidth]{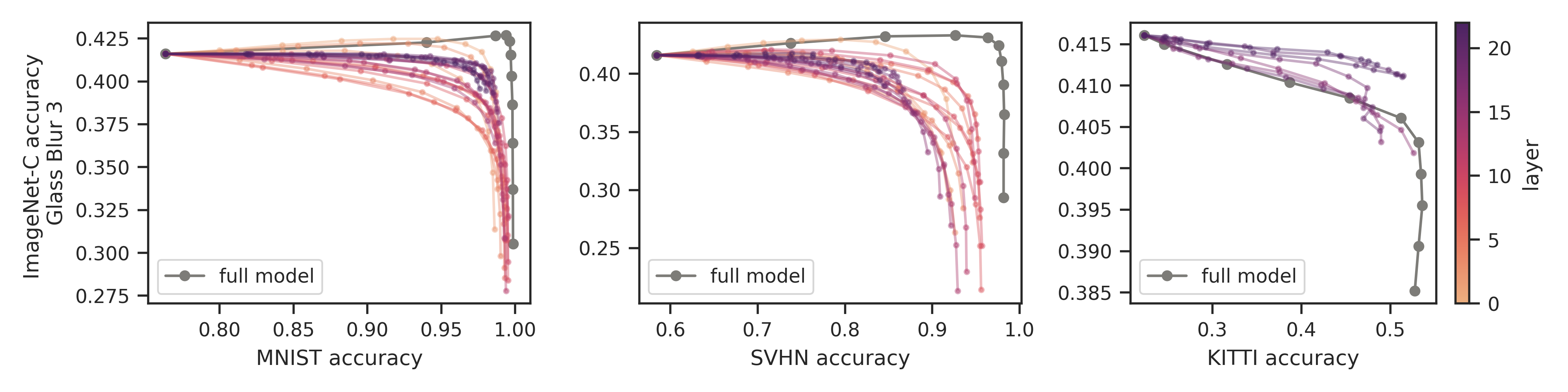}
  \label{fig:1-L-imagenetR}
\end{subfigure}
   \caption{(continued below)}\label{fig:1-L-imagenetC-vs-task-breakdown}
 \end{center}
\end{figure}
\begin{figure}\ContinuedFloat
\begin{center}
\begin{subfigure}[b]{0.95\textwidth}
   \includegraphics[width=1\linewidth]{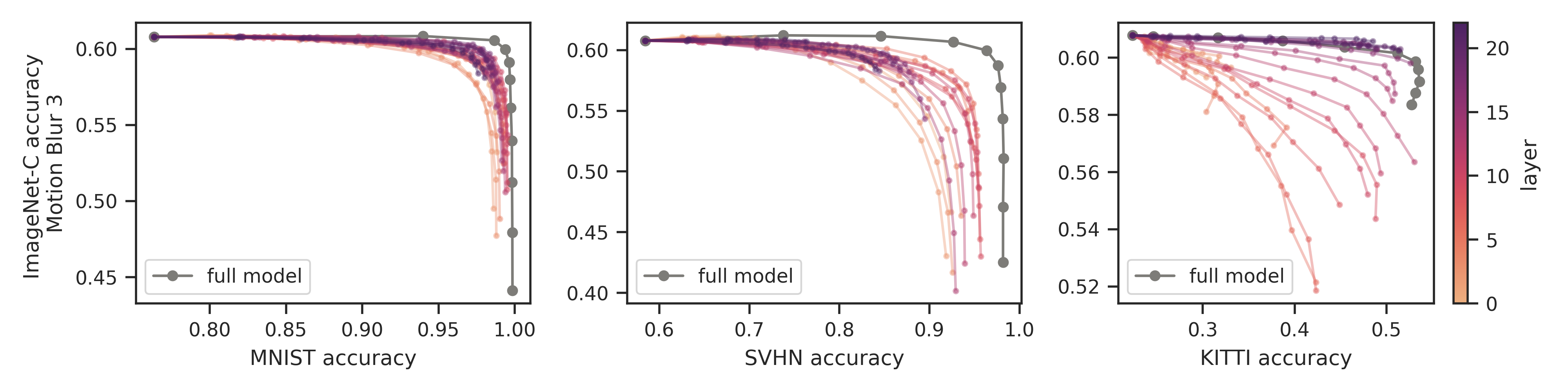}
  \label{fig:1-L-imagenetR}
\end{subfigure}
\begin{subfigure}[b]{0.95\textwidth}
   \includegraphics[width=1\linewidth]{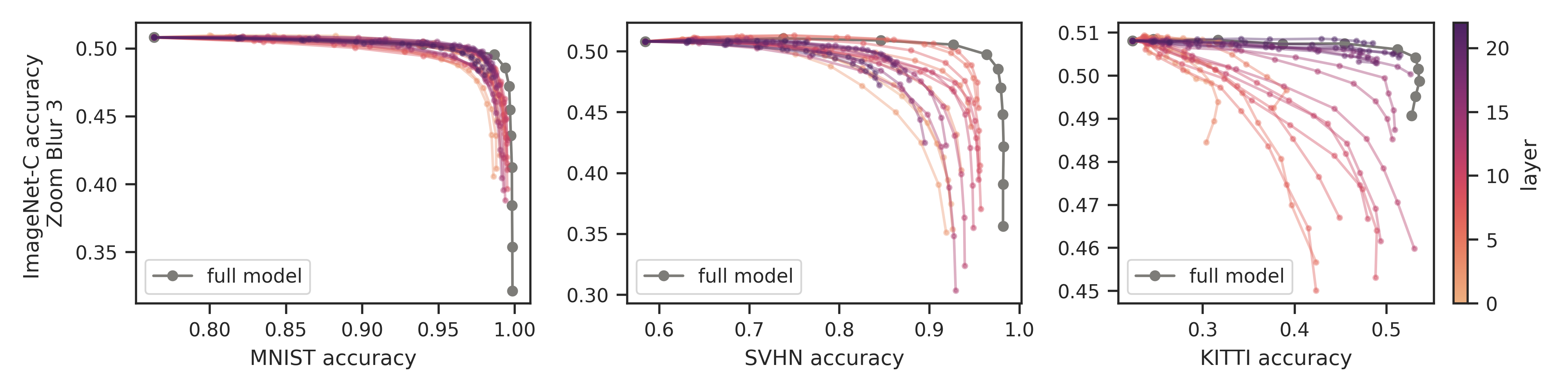}
  \label{fig:1-L-imagenetR}
\end{subfigure}
\begin{subfigure}[b]{0.95\textwidth}
   \includegraphics[width=1\linewidth]{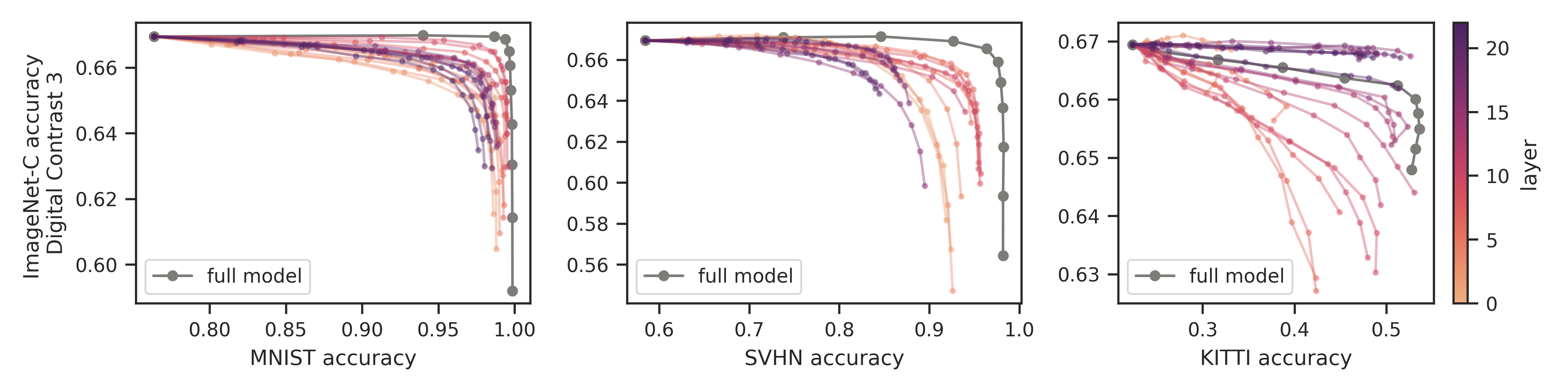}
  \label{fig:1-L-imagenetR}
\end{subfigure}
\begin{subfigure}[b]{0.95\textwidth}
   \includegraphics[width=1\linewidth]{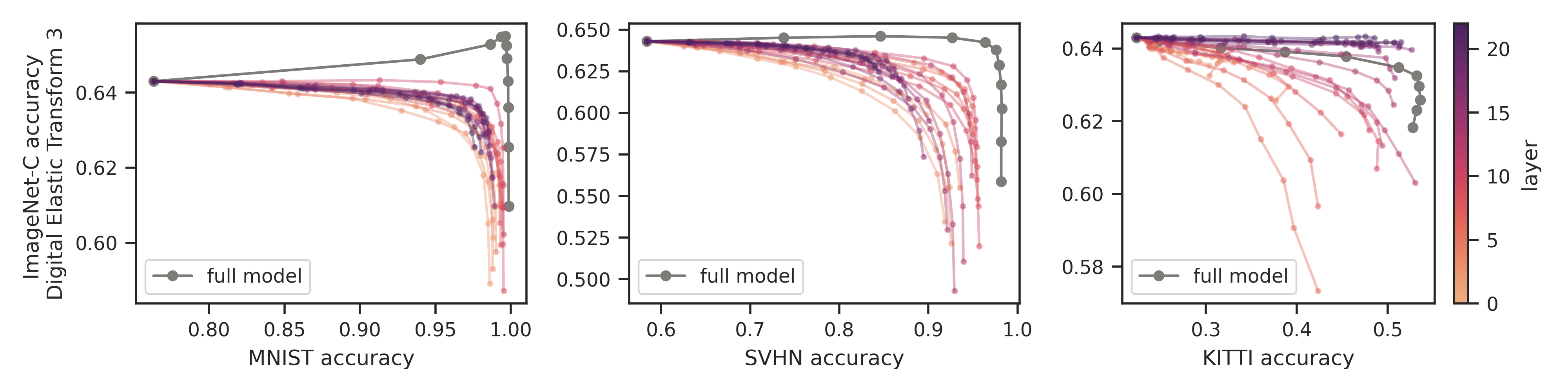}
  \label{fig:1-L-imagenetR}
\end{subfigure}
\begin{subfigure}[b]{0.95\textwidth}
   \includegraphics[width=1\linewidth]{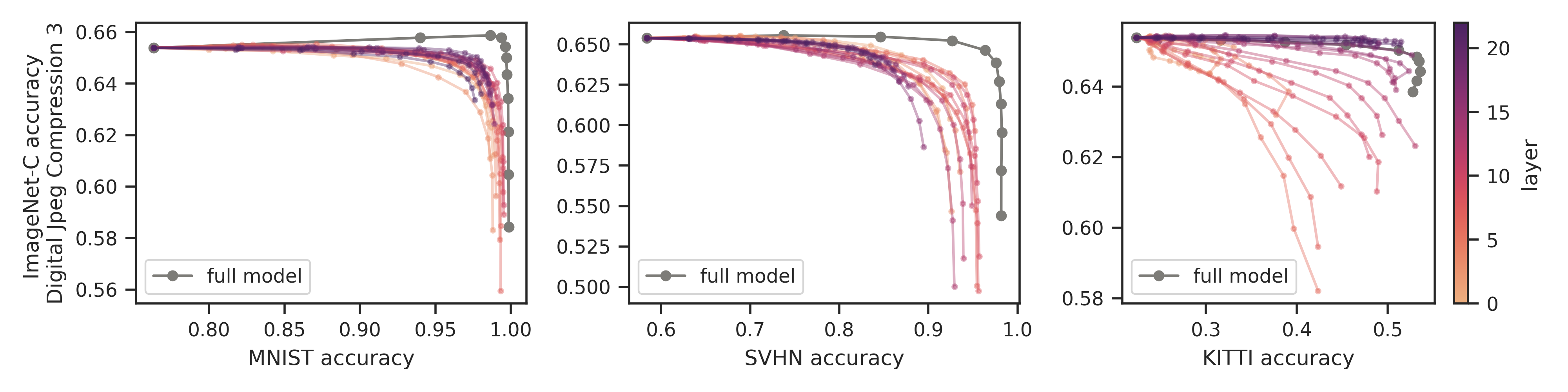}
  \label{fig:1-L-imagenetR}
\end{subfigure}
\begin{subfigure}[b]{0.95\textwidth}
   \includegraphics[width=1\linewidth]{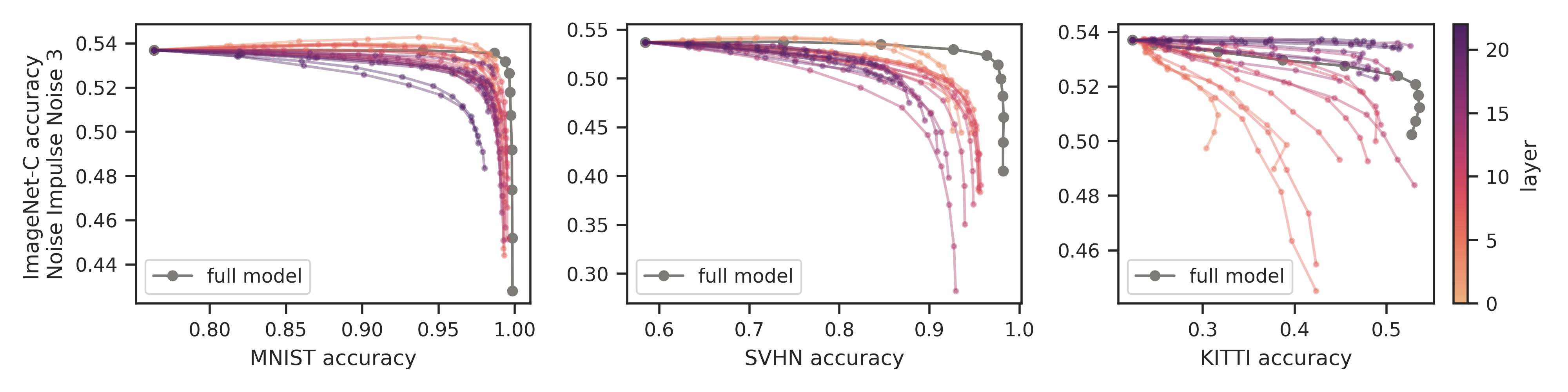}
  \label{fig:1-L-imagenetR}
\end{subfigure}
   \caption{(continued below)}\label{fig:1-L-imagenetC-vs-task-breakdown}
 \end{center}
\end{figure}

\begin{figure}\ContinuedFloat
\begin{center}

\begin{subfigure}[b]{0.95\textwidth}
   \includegraphics[width=1\linewidth]{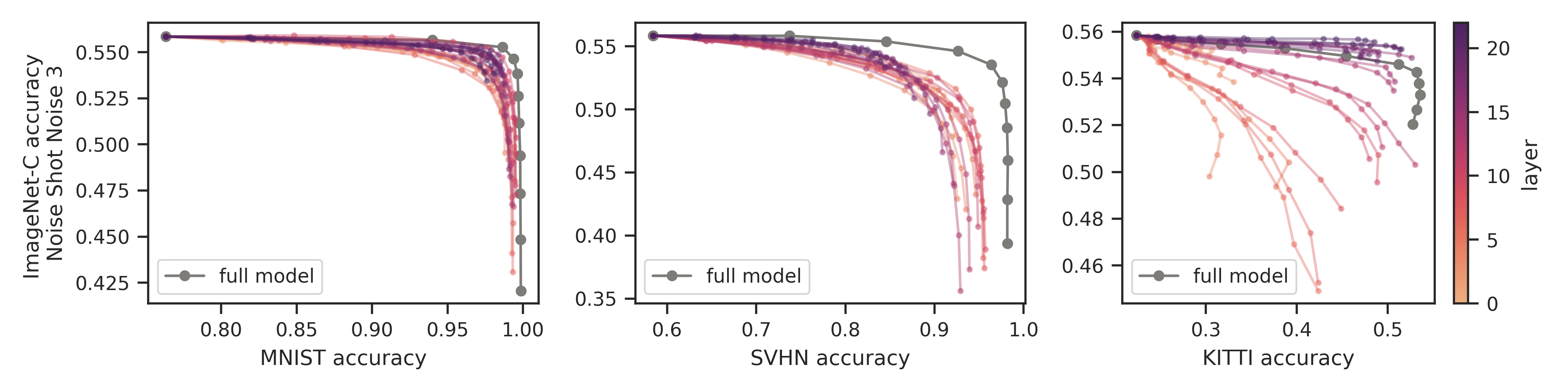}
  \label{fig:1-L-imagenetR}
\end{subfigure}
\begin{subfigure}[b]{0.95\textwidth}
   \includegraphics[width=1\linewidth]{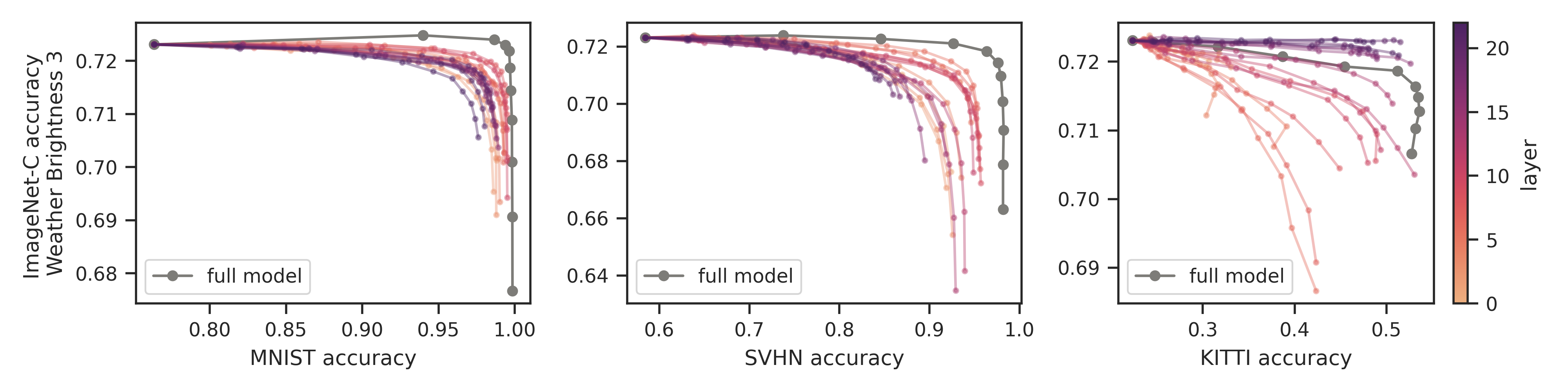}
  \label{fig:1-L-imagenetR}
\end{subfigure}
\begin{subfigure}[b]{0.95\textwidth}
   \includegraphics[width=1\linewidth]{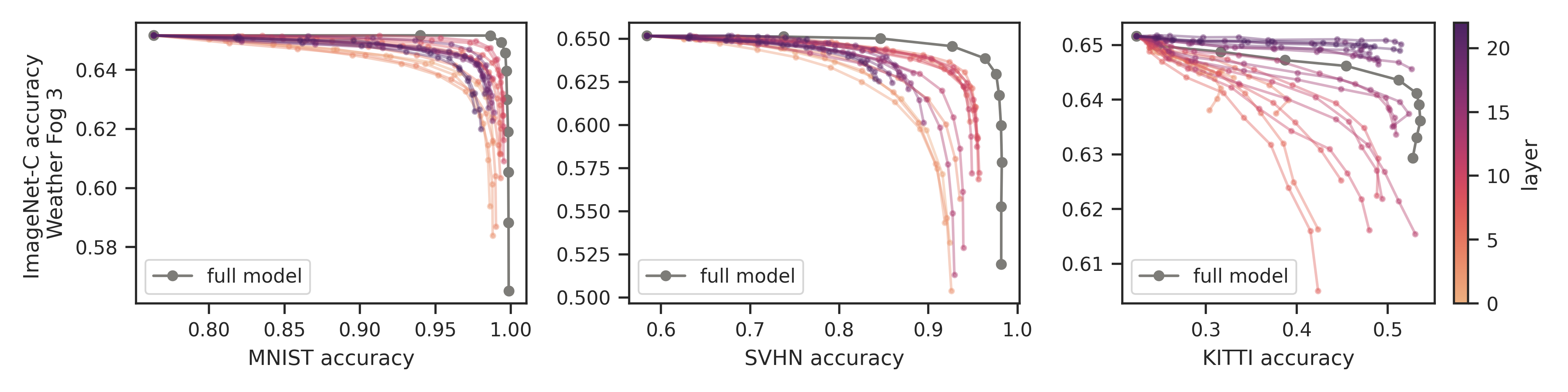}
  \label{fig:1-L-imagenetR}
\end{subfigure}
\begin{subfigure}[b]{0.95\textwidth}
   \includegraphics[width=1\linewidth]{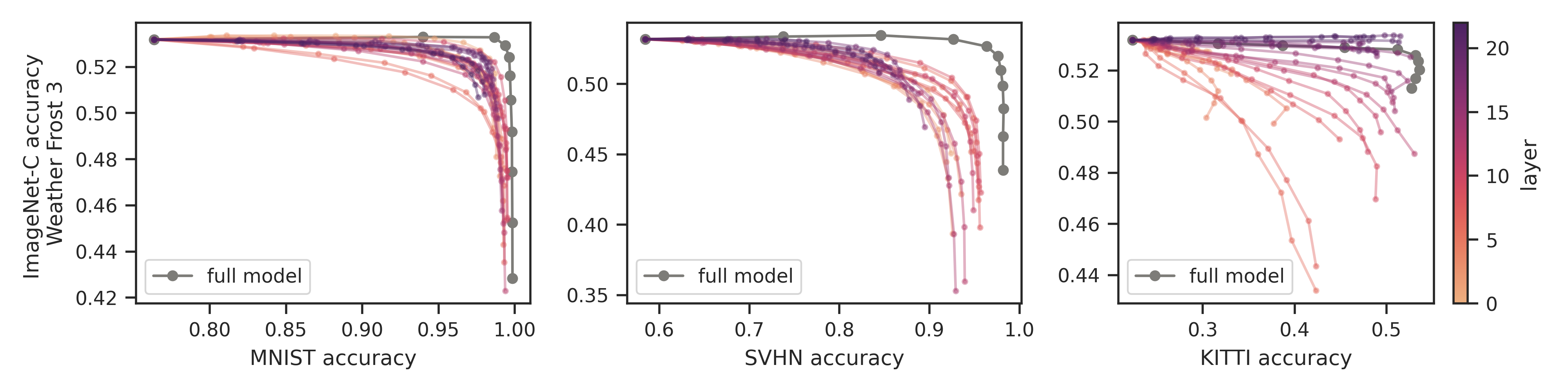}
  \label{fig:1-L-imagenetR}
\end{subfigure}
\begin{subfigure}[b]{0.95\textwidth}
   \includegraphics[width=1\linewidth]{plots/all-datasets-INC-vs-taskresults_C_weather_snow_3.png}
  \label{fig:1-L-imagenetR}
\end{subfigure}
   \caption{Breakdown of ImageNet-C corruption accuracies vs task accuracy, for a corruption strength of 3, aggregated by \ref{fig:1-L-imagenetC}. Computed along line segments \((1-\alpha) W + \alpha W' \) between original and edited weights \(W \) and \(W'\) of CLIP models edited on the given task.}\label{fig:1-L-imagenetC-vs-task-breakdown}
 \end{center}
\end{figure}

\end{document}